\documentclass[a4paper,fleqn]{cas-dc}

\usepackage[numbers]{natbib}
\usepackage{stfloats}
\usepackage{tabularx}
\usepackage{stfloats}
\usepackage{placeins}
\usepackage{cuted}
\usepackage{caption}
\usepackage{hyperref}

\def\tsc#1{\csdef{#1}{\textsc{\lowercase{#1}}\xspace}}
\tsc{WGM}
\tsc{QE}
\tsc{EP}
\tsc{PMS}
\tsc{BEC}
\tsc{DE}

\begin{document}
\FloatBarrier

\let\WriteBookmarks\relax
\def\floatpagepagefraction{1}
\def\textpagefraction{.001}
\shorttitle{A Comprehensive Survey of Medical Image Segmentation: 
Challenges, Benchmarks, and Beyond}
\shortauthors{P.Zhu, X.Zhang, K.Zhang et~al.}

\title [mode = title]{A Comprehensive Survey of Medical Image Segmentation: 
Challenges, Benchmarks, and Beyond}                      

\author[1]{Pengyu Zhu}[orcid=0009-0000-9491-1232]
\fnmark[1]
\ead{pyzhu@ncepu.edu.cn}
\credit{Investigation, Formal analysis, Data curation,Visualization, Writing-original draft, Writing-review \& editing}

\affiliation[1]{organization={School of Control and Computer Engineering, North China Electric Power University},city={Beijing},postcode={102206},country={China}}

\author[1,2]{Xiaojing Zhang}
\fnmark[1]
\ead{xiaojingzhang@ncepu.edu.cn}
\credit{Data curation, Formal analysis, Investigation,Visualization, Writing-original draft, Writing-review \& editing}

\affiliation[2]{organization={SPIC Digital Technology Co., Ltd},city={Beijing},postcode={102209},country={China}}

\author[3]{Kunbo Zhang}
\ead{kunbo.zhang@ia.ac.cn}
\credit{Writing-review \& editing}

\affiliation[3]{organization={Institute of Automation, Chinese Academy of Sciences},
                city={Beijing},
                postcode={100190},
                country={China}}

\author[4]{Chunyan Zhang}
\ead{xchunxchun@163.com}
\credit{Investigation, Funding acquisition, Writing-review \& editing}

\affiliation[4]{organization={Department 6 of Health Care, Second Medical Center, People's Liberation Army General Hospital},
                city={Beijing},
                postcode={100037},
                country={China}}

\author[1]{Zhenyu Wang}
\cormark[1]
\ead{zywang@ncepu.edu.cn}
\credit{Investigation, Funding acquisition, Writing-review \& editing}

\fntext[1]{These authors contributed equally to this work.}
\cortext[cor1]{Corresponding author}

\begin{abstract}
Medical image segmentation plays a critical role in clinical diagnostics, treatment planning, disease monitoring, and neurological disorder identification.
This article presents a comprehensive review of its systematic development, covering widely used public datasets, representative methods built on the U-Net, Transformer, and SAM architectures, and key evaluation metrics with their differences, followed by an analysis of major challenges from multiple perspectives.
Unlike surveys that focus on a single model family or a specific clinical application, this review organizes U-Net-, Transformer-, and SAM-based methods within a unified analytical framework, with a particular focus on their effectiveness in improving segmentation accuracy and efficiency.
This work aims to guide future research and support clinical translation of medical image segmentation, with all related resources publicly available in our GitHub repository \href{https://github.com/andrew-pengyu/Awsome_MedSeg/tree/main}{Awsome\_MedSeg}.
\end{abstract}




\begin{keywords}
Medical Image Segmentation \sep U-Net \sep Transformer \sep SAM \sep Challenge
\end{keywords}

\maketitle

\section{Introduction}

Medical image segmentation is an important task in the field of medical image analysis, with the focus of accurately identifying and delineating the target structures within the medical modalities such as CT and MRI, so that it can be efficiently distinguished from the background \cite{36D-Personal,60TIP1,61TIP_acdc}. 
It supports important clinical tasks such as disease diagnosis, treatment planning, image-guided surgery, and monitoring of disease progression. Being a central part of the whole medical imaging pipeline, segmentation generally accepts preprocessed images as input and returns pixel- or voxel-level labels which define anatomical or pathological regions of interest. These maps offer important structural information for subsequent tasks like classification, registration and quantitative analysis.

Conventional segmentation techniques rely on hand-designed features and heuristic algorithms, which require substantial domain knowledge and manual intervention \cite{41by1}. However, they have a hard time dealing with complex anatomical appearance variance, noise or low quality images \cite{42by2}, and fail to generalize to other modalities or even patient population. The revolution in the field is led by deep learning, specifically U-Net, Transformer, and SAM architectures. Through learning hierarchical image features, such models greatly increase both segmentation performance and robustness, minimizing the reliance on expert and manual intervention and achieving new levels of degree of automation. 

Recent models achieve near real-time inference, producing results within seconds per volume, and reach expert-level performance in tasks such as tumor delineation, organ and lesion segmentation.
However, limitations persist such as poor cross-institution generalization, modality-dependent performance, lack of annotation, and suboptimal performance on complex or rare pathologies.
Additionally, there is extensive variation in the segmentation task over different imaging modalities (CT for dense structures and MRI for soft tissue), dimensions (2D slices and 3D volumes), clinical application scenarios (real-time surgical guidance and retrospective analysis). 
Design of models has to consider the spatial context, computational efficiency and anatomical variations, making multi-modality and adaptability become more and more important.

In recent years, a growing number of review articles have focused on medical image segmentation.
For instance, Zhang et al.\cite{38re1} center their discussion around the recently released SAM model and its variants, providing valuable insights into the latest developments in foundation models. Yao et al.\cite{39re2} and M.E.Rayed et al.\cite{40re3} offer reviews from specific methodological and application-oriented perspectives.
Although several recent surveys have reviewed medical image segmentation, foundation models, or specific application scenarios, a unified discussion that explicitly contrasts U-Net-, Transformer-, and SAM-based methods from the perspectives of data dependency, architectural bias, evaluation practice, and clinical deployment remains valuable.

This article provides a comprehensive survey of the area of medical image segmentation using deep learning, concentrating on four central elements:

\textbf{Mainstream Datasets}: We introduce a few small datasets like BraTS and LIDC-IDRI, that focus on particular tissues or organs, as well as some large-scale ones such as Synapse Multi-Organ CT, and explain their properties. Although these datasets have greatly advanced medical image analysis, most remain limited in scale, population diversity, and anatomical coverage, often restricted to single-organ or single-modality tasks. In addition, inconsistent annotation quality hampers reproducibility and generalization. Future work should emphasize developing large-scale, multi-modal, and multi-organ datasets with standardized, high-quality annotations.

\textbf{Evolution of Models}:
We follow the evolution of segmentation architectures from early convolutional networks to state-of-the-art foundation models. The representative architectures are classified into U-Net-based, Transformer-based, and SAM-derived models, indicating a new trend that the design of the model is transforming from task-oriented models to universal and general models according to specific demands. This transformation is associated with an increasing focus on versatility, domain adaptability, and clinical robustness. Future work will investigate hybrid models combining prior anatomical knowledge, multi-modality data, and text-prompt interactions in an end-to-end effective framework.

\textbf{Key Evaluation Metrics}: We present an extensive summary of the main evaluation measures, including their definition, calculation, and interpretability. This is largely because of the various clinical and task-dependent needs, where establishing a common measure is challenging. Rather than seeking a one-size-fits-all standard, future work should explore the context-aware metric selection, composite or task-adaptive schemes, and aim to be more aligned with clinical relevance and uncertainty estimation for fair and actionable assessments.

\textbf{Principal Challenges}: We gather all the problems that exist in dataset distribution, algorithm development, and clinical application. Problems like distributional mismatch, lack of interpretability, and poor integration with diagnostic protocols make it very hard to use in real life. To solve these problems, engineers and doctors need to work together more closely, and the community needs to set up clinically validated benchmarks and validation protocols that take the situation into account.

\section{Mainstream Datasets}
With the advancement of deep learning, a variety of medical image segmentation datasets are widely adopted to train and test models. This article introduces the current mainstream medical image segmentation datasets depicted in Table I, explores their specific characteristics and highlights the pathological tissues and organs they target, as illustrated in Fig. 1.

\begin{table}[htbp]
\centering
\scriptsize 
\caption{Summary of Public Medical Image Segmentation Datasets}
\renewcommand{\arraystretch}{1.1} 
\begin{tabularx}{\linewidth}{@{}c| c c c@{}}
\Xhline{1pt}
\textbf{Dataset} & \textbf{Modality} & \textbf{\#Cases} & \textbf{Target Organ/Disease} \\
\Xhline{1pt}
BraTS \cite{49BraTS} & MRI & 259 & Brain Tumor \\
\midrule
LIDC-IDRI \cite{54LIDC-IDRI} & CT & 1018 & Lung Nodules \\
\midrule
ACDC \cite{55ACDC} & cine-MRI & 150 & Heart Structures \\
\midrule
LA \cite{56LA} & GE-MRI & 100 & Left Atrium \\
\midrule
LiTS \cite{57lits} & CT & 200 & Liver Tumor \\
\midrule
KiTS \cite{58kits} & CT & 210 & Kidney Tumor \\
\midrule
Pancreas-CT \cite{1BCP} & CT & 82 & Pancreas \\
\midrule
PROMISE12 \cite{33V-Net} & MRI & 50 & Prostate \\
\midrule
\multirow{4}{*}{QUBIQ \cite{59qubiq}} 
 & MRI & 55 & Prostate \\
 & MRI & 39 & Brain Growth \\
 & Multi-modal MRI & 32 & Brain Tumor \\
 & CT & 24 & Kidney \\
\midrule
Synapse \cite{25SAMed} & CT & 30 & 8 Abdominal Organs \\
\Xhline{1pt}
\end{tabularx}
\label{tab:medseg-datasets}
\end{table}

\subsection{The BraTS Dataset}

The BraTS (Brain Tumor Image Segmentation Benchmark) dataset \cite{49BraTS} is a high-quality multi-modal MRI dataset dedicated to brain tumor segmentation studies, comprising T1-weighted, T1GD-enhanced, T2-weighted and FLAIR images\cite{50flair,62TIP_brats}.
Each image is manually annotated by professional doctors to cover the three major regions of enhanced tumor, edema and core, enabling high-precision segmentation and the evaluation of algorithm performance \cite{8CM-Pixel}. 
This dataset is revised and refined annually, and the scale as well as the quality of annotation are consistently increasing.

\subsection{The LIDC-IDRI Dataset}

The LIDC-IDRI dataset \cite{54LIDC-IDRI} consists of chest CT scan images of 1018 patients, stored in DICOM format \cite{51dicom} to standardize and support the data format.
This dataset encompasses 1018 research cases, each image annotated by four experienced radiologists across two stages. In the first stage, the radiologists independently diagnose and annotate nodules that are 3 mm or larger as well as those smaller than 3 mm, followed by a second stage involving re-evaluation to confirm the final diagnosis. 
This multi-stage, multi-expert annotation approach captures the variability in clinical annotation practices \cite{36D-Personal}, aiding models in handling uncertainty and diversity.

\subsection{The ACDC Dataset}
The ACDC dataset\cite{55ACDC} focuses on segmenting the left ventricle, right ventricle and myocardium in dynamic magnetic resonance imaging of the heart, covering 150 cases across 5 subcategories: normal, myocardial infarction, dilated cardiomyopathy, hypertrophic cardiomyopathy and right ventricular abnormalities, with 30 cases per category \cite{1BCP,4AutoSAM,9CoraNet,28SS-Net,37Tyche,46show2025,61TIP_acdc}. 
Each of the cases has the 4D NIfTI format images of the cardiac cycle, consisting of end-diastolic and end-systolic frames. The dataset is divided into 100 training cases and 50 testing cases, with 20 and 10 cases per subcategory in the training and testing sets, respectively.

\begin{figure*}[!t]
\centering
\footnotesize
\includegraphics[width=0.75\textwidth]{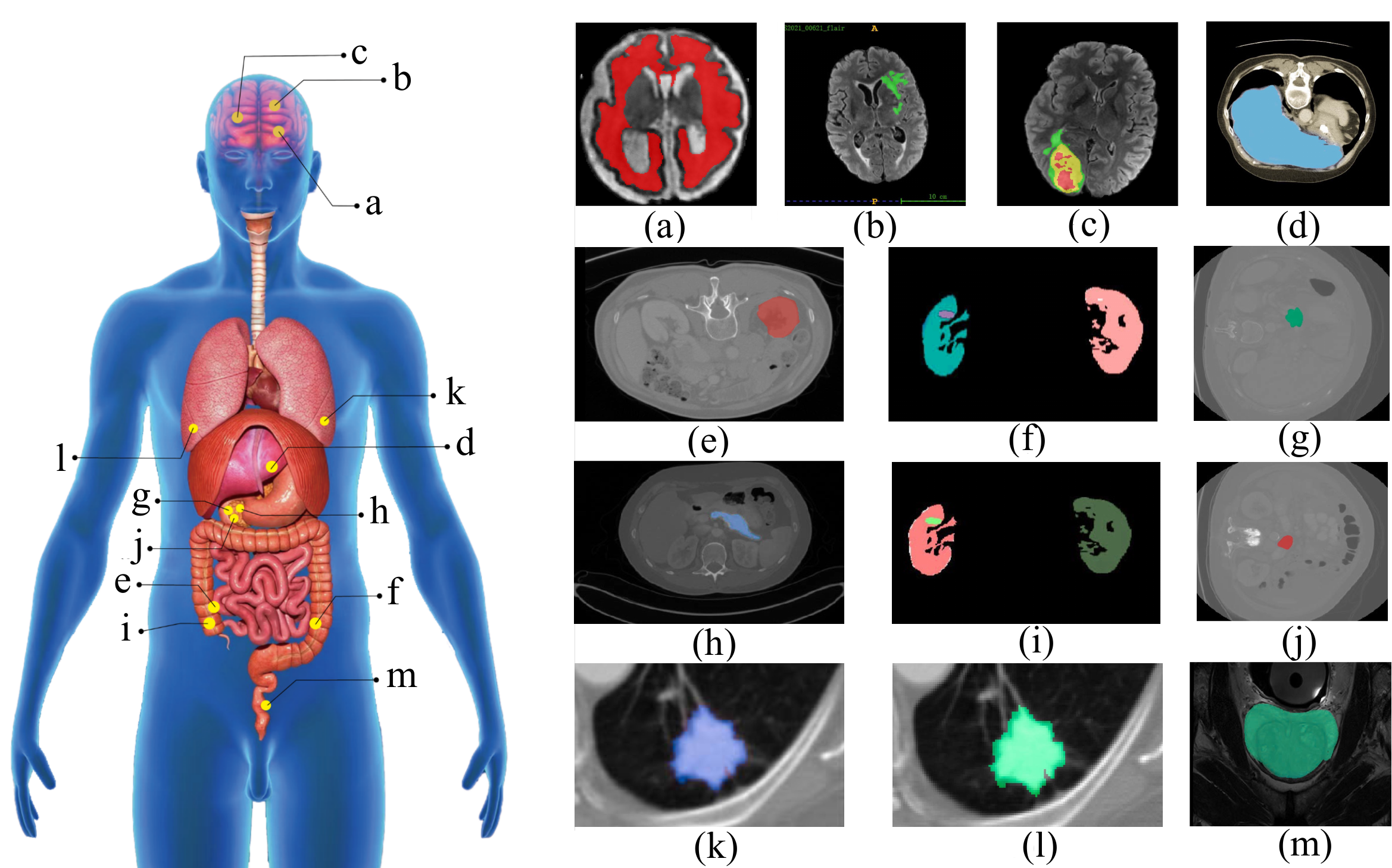}
\caption{Current mainstream datasets and their corresponding human organ distribution maps.(a) shows brain growth, and (b) shows brain tumor images from the QUBIQ dataset\cite{59qubiq}, respectively. (c) Brain tumor segmentation images from the BraTS 2021 dataset \cite{49BraTS}. (d) Liver segmentation images from the LiTS17 dataset\cite{57lits}. (e) Kidney segmentation images from the QUBIQ dataset. (j) and (h) are pancreatic islet images from the Pancras CT\cite{1BCP} and QUBIQ datasets, respectively. (f) and (i) show left and right kidney segmentation images from the KiTS23 dataset\cite{58kits}, respectively. (g) Pancreatic islet injury images from the QUBIQ dataset. (k) and (l) show lung nodule segmentation images from the LIDC-IDRI dataset\cite{54LIDC-IDRI}. (m) Prostate segmentation images from the PROMISE12 dataset\cite{33V-Net}.}
\label{fig1}
\end{figure*}

\subsection{The LA Dataset}
The LA (Left Atrium) dataset\cite{56LA} is a high-quality medical imaging dataset designed specifically for left atrial segmentation tasks, including 100 3D gadolinium-enhanced magnetic resonance imaging (GE-MRI) scan images, each annotated with detailed left atrium labels by professional cardiologists, guaranteeing the high accuracy, consistency and credibility of the data \cite{13MC-Net,28SS-Net,30UA-MT}. 
The LA dataset is constructed by high-resolution and high-contrast images which make heart structure and lesion areas very clear. There are various sections and angles of the left atrium in each picture of the dataset, which can support researchers with enough cardiac imaging information.
Imaging processing (e.g. normalization and resampling) is necessary to adapt the LA dataset for research across the algorithms.

\subsection{The LiTS Dataset}
The LiTS (Liver Tumor Segmentation) dataset \cite{57lits} is a medical imaging dataset made just for liver and liver tumor segmentation. 
It includes 200 liver CT images, which are carefully annotated by an experienced radiologist with the manual segmentation annotations for the liver and liver tumors. These labels are completely accurate and consistent, so researchers can use them to train and test their models with high-quality data. 
The data is fairly balanced, containing a training set of 130 images and a test set of 70 images so that researchers can systematically develop and assess their segmentation models.

\subsection{The KiTS Dataset}
The KiTS (Kidney and Kidney Tumor Segmentation Challenge) dataset \cite{58kits} is a collection of real kidney CT scan images from all over the world that focuses on kidney tumor segmentation. It holds a large collection of CT images of the kidney in DICOM format, which can be used and analyzed by medical imaging software. Every two years, this dataset gets new versions, such as KiTS 2021 and KiTS 2023. It also provides matching annotation files for training and testing algorithms.

\subsection{The Pancreas-CT Dataset}
The Pancreas-CT dataset \cite{1BCP,9CoraNet,12DTC}, which is dedicated to pancreatic segmentation, comprises 82 abdominal CT scans performed by the National Institutes of Health Clinical Center from 80 subjects including healthy kidney donors and patients without severe abdominal or pancreatic lesions.
Participants' age varied from 18 to 76, with a mean of 46.8. The scans were acquired by Philips and Siemens MDCT scanners with the resolution of $512 \times 512$ and the slice thickness between 1.5 and 2.5 mm.
The pancreatic partition was manually performed by a medical student and validated and corrected by experienced radiologists in order to ensure labeling precision.

\subsection{The PROMISE12 Dataset}
The PROMISE12 (Prostate MR Image Segmentation 2012) dataset \cite{33V-Net,47steady2025} is a publicly available dataset which is tailored for prostate MR image segmentation.
This dataset comprises 50 3D cross-sectional T2-weighted MR images with a resolution of $512 \times 512$ pixels, along with manual segmentation annotations of the prostate gland for each image. 
The images in this dataset originate from multiple medical centers and have been scanned using various imaging protocols, enhancing the data's diversity and representativeness and aiding in the assessment of related models' generalization capabilities. 
In the classical model research, there is only one train-test split on the PROMISE12 dataset, which is divided into 40 training cases and 10 test cases.
A small portion of the training set (e.g., 3 examples) is used as labeled data for supervised learning experiments and the others are treated as unlabeled data for semi-supervised learning experiments. 

\subsection{The QUBIQ Dataset}
The QUBIQ (Quantification of Uncertainties in Biomedical Image Quantification) dataset \cite{59qubiq} is generated for quantifying and handling the discrepancies in the segmentation results of medical images.
The novelty of this dataset is contributing multiple annotations across diverse domains, thus establishing a rich data backbone for annotation inconsistency and uncertainty analysis.
This dataset contains seven different segmentation tasks: two prostate segmentation tasks (MRI), one brain growth segmentation task (MRI), three brain tumor segmentation tasks (multi-modal MRI), and one kidney segmentation task (CT).
This corpus includes 39 cases with 7 types of brain growth annotations, 32 cases with 3 types of tumor growth annotations on brain, 24 cases with 3 types of kidney annotations, and 55 cases with 6 types of prostate annotations. Of these, 48 prostate cases, 34 brain growth cases, 28 brain tumor cases, and 20 kidney cases were used for training, with the remaining cases used for testing.

\subsection{The Synapse Multi-Organ CT Dataset}
The Synapse dataset\cite{25SAMed,64TIP_synapse} comprises a total of 3779 axial contrast-enhanced abdominal CT images. The training set contains 2212 axial slices. The dataset is divided into 18 training cases and 12 testing cases. Each CT scan in the Synapse dataset includes between 85 and 198 slices, with a resolution of $512 \times 512$ pixels during low-sample training and $224 \times 224$ pixels during fully supervised training. The dataset includes annotations for 8 major abdominal organs, including the aorta, gallbladder, spleen, left kidney, right kidney, liver, pancreas, and stomach. All annotations were done manually by experienced experts and were subjected to rigorous quality control and validation to achieve an acceptable level of accuracy and consistency.

\section{Evolution of Models}
Various deep learning structures have been presented for medical image segmentation in the past decades. As illustrated in Fig. 2, the progress of these models has taken a fairly clear evolutionary path from CNN (e.g., U-Net) to Transformer-based models and most recently to foundation models such as SAM (Segment Anything Model), as shown in Fig. 3. Beyond these mainstream backbones, the new studies have adopted more advanced approaches such as knowledge distillation and conditional normalizing flows, providing more diversity in the space of segmentation models. These emerging architectures not only improve performance but also represent different views for benchmarking future models.

\begin{figure*}[!t]
\centering
\footnotesize
\includegraphics[width=0.85\textwidth]{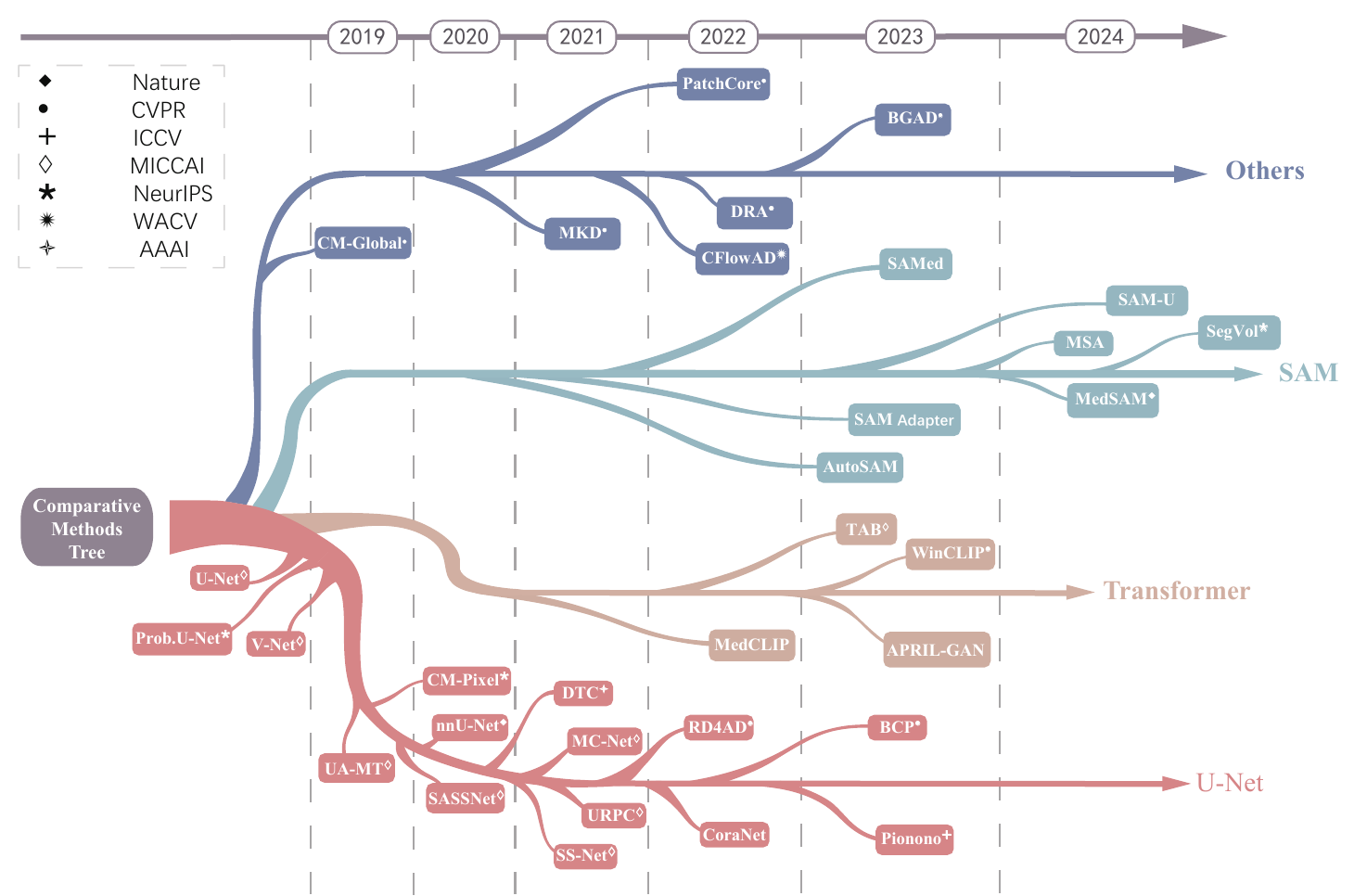}
\caption{Comparison method tree diagram. In recent years, advanced algorithms based on U-Net, Transformer, SAM and other methods have been employed to evaluate and compare model performance.}
\label{fig2}
\end{figure*}

\subsection{U-Net Based Methods}

U-Net\cite{31U-Net}, a fully convolutional network for biomedical image segmentation, represents a hallmark in the field due to its high efficiency and capability to learn from few annotated samples. The symmetrical U-shape design of U-Net, comprising contraction and expansion paths linked by skip connections, allows aggregation of both global context and local details effectively. During the past time, U-Net has been extended to different data scenarios which has led to the development of supervised, semi-supervised, and unsupervised U-Net-based architectures, making for its wider application and enhancing the robustness in complex segmentation tasks on medical data. 

Supervised learning approaches depend solely on the labeled data, and lots of the variations of U-Net also belong to this type. In order to deal with 3D volumetric data and class imbalance, V-Net \cite{33V-Net} is an extension of U-Net to three dimension, with a Dice-based loss function which is modified for the imbalanced segmentation tasks. Focusing on the problem of manual design and dataset-dependent tuning, nnU-Net \cite{18nnU-Net} devises a self-configuring framework that can automatically adjust its architecture and training regime according to various datasets, exhibiting robust performance across heterogeneous tasks with least human intervention. Meanwhile, to cope with uncertainty and inter-observer variability in clinical settings, Pionono \cite{20pionono} combines probabilistic modeling with the U-Net architecture. It uses ResNet34-based feature extraction and Monte Carlo sampling for prediction uncertainty estimation. These models demonstrate that the U-Net approach can be tailored to enhance generalization, automation, and robustness in real-world medical scenarios.

Due to high annotation costs in medical imaging, semi-supervised learning is becoming more crucial. These techniques can take advantage of a limited number of labeled samples, combined with large sets of unlabeled images to boost performance. To tackle ambiguous medical image segmentation, Probabilistic U-Net \cite{21Prob.U-Net} integrates a Conditional Variational Autoencoder (CVAE) to produce multiple viable segmentation hypotheses. Uncertainty modeling and consistency learning are also popular: UA-MT \cite{30UA-MT} utilizes teacher-student framework incorporating uncertainty-aware consistency, whereas MC-Net \cite{13MC-Net} employs dual-decoder uncertainty learning alongside cyclic pseudo-labeling. To mitigate the impact of noisy or inconsistent annotations, CM-Pixel \cite{8CM-Pixel} and CoraNet \cite{9CoraNet} use coupled networks and self-training with uncertainty filtering, and URPC \cite{32URPC} applies uncertainty correction to further refine pseudo-labels. To fix the problem of limited feature representation, multi-task models such as SASSNet \cite{27SASSNet} and DTC \cite{12DTC} integrate tasks like signed distance prediction and global structure modeling to reinforce semantic learning. Additional enhancements come from architectural novelties: SS-Net \cite{28SS-Net} brings in adversarial perturbations and prototype guidance, while BCP \cite{1BCP} alleviates domain gaps by copy-paste augmentation. These approaches reflect the practice of combining U-Net with auxiliary techniques such as uncertainty estimation, consistency learning, and task regularization to exploit learning from scarce annotated data.

In the absence of annotations, unsupervised techniques are a potential alternative. RD4AD \cite{22RD4AD} proposes a U-Net-based architecture for anomaly detection through reverse distillation and bottleneck feature learning. It demonstrates that such models are great at finding anomalous regions, without the need for pixel-level labels, and has a promising potential for applications such as screening or diagnosis aid. 

In the three above categories, U-Net and its derivatives target the same objective of boosting the accuracy and robustness of segmentation within the architectural and learning techniques.  Supervised methods improve accuracy by refining architectures and loss functions, semi-supervised methods aim to learn effectively with limited data, and unsupervised methods aim for more general applicability where annotations cannot be acquired. Together, they demonstrate the versatility of U-Net as a building block in medical image segmentation and offer a diverse set of design choices to future work.

\subsection{Transformer Based Methods}
{The Transformer architecture has shown strong capacity for modeling global dependencies and complex relations, particularly in domains such as medical image segmentation, image-text matching, and few-shot learning. 

For annotation variability and segmentation precision, TAB \cite{29TAB} introduces a multivariate normal distribution in the Transformer architecture to describe annotator bias, which can capture the annotator preferences and output smooth and continuous segmentation boundaries.
To tackle data scarcity and modality alignment, MedCLIP\cite{14Medclip} adopts a vision-text contrastive learning framework, which is boosted using extracted medical knowledge. By projecting both visual and textual sources into a single embedding space, it improves image-text matching performance and demonstrates favorable data efficiency and generalization ability.
To address the problem of limited supervision in few-shot scenarios, APRIL-GAN\cite{3APRIL-GAN} utilizes the CLIP model and adds a memory banking mechanism to enable efficient abnormality detection and segmentation in industrial inspection task trained with a small number of labeled samples. These models showcase excellent potential of the Transformer to capture global dependencies and learn complex relationships, outperforming conventional CNN-based methods by a large margin.
}

\begin{figure*}[!t]
\centering
\includegraphics[width=0.85\textwidth]{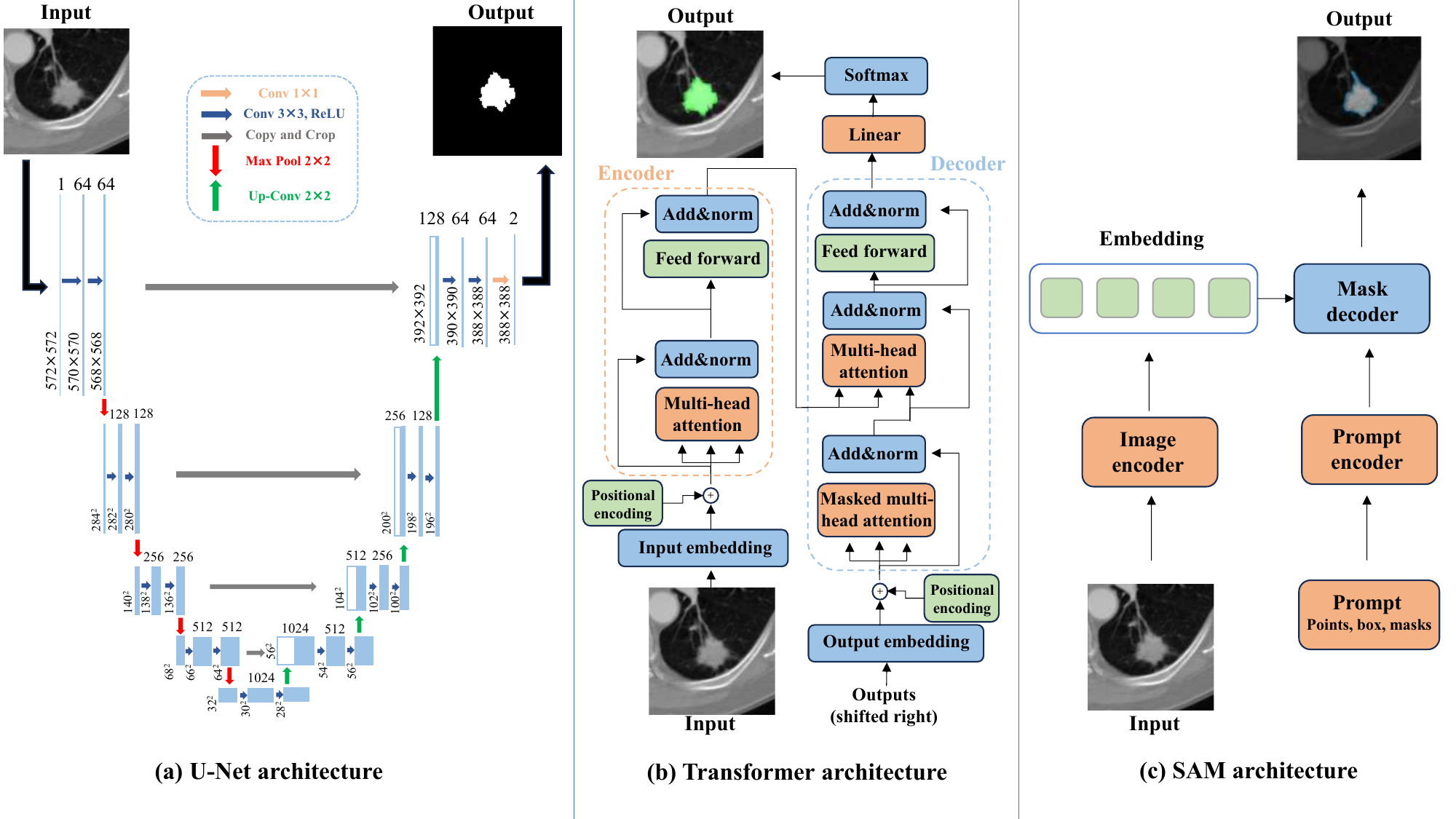}
\caption{Comparison of U-Net, Transformer, and SAM model architectures. (a) depicts the U-Net model framework, (b) illustrates the Transformer model architecture, and (c) shows the Segment Anything (SAM) model architecture.}
\label{fig3}
\end{figure*}

\subsection{SAM Based Methods}

While the Segment Anything Model (SAM)\cite{23SAM} is inherently based on the Vision Transformer (ViT), it introduces a new paradigm in image segmentation: prompt-driven, general-purpose foundation models. This paradigm has shown remarkable potential in medical imaging, where annotation is costly and data distribution varies widely. To adapt SAM for medical use, recent works have explored three main directions: domain adaptation to medical data, reducing dependence on prompts and heavy fine-tuning, and enhancing robustness in complex or uncertain scenarios.

To address modality adaptation and medical domain specificity, MedSAM \cite{15MedSAm} fine-tunes large-scale models across diverse modalities, and attains state-of-the-art results on multi-organ and multi-modal segmentation. The SAM Adapter \cite{24SAMAdapter} embeds lightweight domain-specific adapters into SAM, that can be employed to perform efficient fine-tuning for complex segmentation tasks such as organ camouflage or subtle boundaries.
For the demands of lower prompt and parameter dependency, SAMed \cite{25SAMed} adopts low-rank adaptation to reduce the number of trainable parameters, greatly lowering the deployment cost without a compromise on the accuracy, while AutoSAM \cite{4AutoSAM} removes the manual prompt input by integrating an automatic prompt generator, achieving fully automated or low-supervision segmentation.
To enhance robustness and scalability, SAM-U \cite{26SAM-U} incorporates uncertainty estimation, improving its robustness to noise or low-quality imaging. MSA \cite{17MSA} generalizes SAM into 3D imaging through interactive fine-tuning mechanisms, and SegVol \cite{43du2023segvol} further integrates spatial and semantic prompts to support fine-grained, interactive tumor segmentation across volumetric modalities.

In short, these methods can be viewed as the gradual fit of SAM from a universal vision model to a universal medical segmentation backbone. They develop solutions to domain transfer, prompt efficiency, and robust generalization, which can unleash the potential of SAM across a wide variety of clinical applications.

\subsection{Other Methods}

In addition to the prevalent U-Net, Transformer and SAM-based paradigms, several models adopt alternative methodologies such as uncertainty modeling, knowledge distillation, flow-based density estimation and contrastive learning, to cope with domain-specific issues like label noise, annotation scarcity, and open-set anomaly detection. These approaches typically highlight flexibility, interpretability and the practicability in real life medical and industrial applications.

In order to alleviate the influence of noisy or inconsistent annotations, CM-Global \cite{7CM-Global} proposes a regularized loss to learn annotator confusion matrices, which helps improve the quality of labels and the segmentation accuracy of multi-annotator datasets. For interpretable and efficient anomaly detection, MKD \cite{16MKD} adopts multi-resolution feature extraction with knowledge distillation to produce gradient-based explanations with precise localization. For data scarcity and domain shift, DRA \cite{11DRA} is proposed to learn on normal and few anomaly data simultaneously with a multi-head architecture, and is shown to perform well in terms of cross-domain generalization. In the same way, PatchCore \cite{19Patchcore} resolves the cold-start issue by local feature embedding and greedy subsampling, enabling fast and accurate anomaly detection with minimal training. For unsupervised or low-supervision scenarios, CFlow-AD \cite{6CFlow-AD} leverages conditional normalizing flow model to estimate pixel-wise likelihood for real-time, label-free anomaly detection, and BGAD \cite{5BGAD} unites pseudo anomaly generation and boundary-guided contrastive learning to promote generalization to both seen and unseen anomalies. Lastly, for fast, generalizable segmentation across diverse inputs, Iris \cite{46show2025} couples a 3D U-Net and a Transformer backbone with a shallow task encoder, this leads to an efficient multi-class segmentation in a single forward pass.

Taken together, these approaches make the segmentation field more prosperous with various novel architectures and learning paradigms to solve practical limitations such as few labels, unclear boundaries, and real-time computation.

\section{Key Evaluation Metrics}

In medical image segmentation, robust evaluation is critical due to noise and class imbalances and image interpretation ambiguity in manual annotation. In contrast to standard classification approaches which only consider precision, recall and F1-score, segmentation requires a more comprehensive evaluation that describes region overlap, classification quality, and boundary accuracy to provide a more complete picture of model performance.
Such metrics not only establish fair and interpretable benchmarks across datasets and applications, but also help algorithm development from technical and clinical standpoints.

\subsection{Overlap-Based Metrics}

\subsubsection{DSC}
{
The Dice Similarity Coefficient(DSC) is frequently employed in medical image segmentation tasks \cite{1BCP,2ADVENT,13MC-Net,15MedSAm,17MSA,25SAMed,26SAM-U,27SASSNet,30UA-MT,32URPC,63TIP_DSC} to quantify the overlap between segmentation outcomes and expert annotations. Its definition is given by the following formula:
\begin{equation}\label{eqn-1} 
 Dice=\frac{2\mid A\cap B\mid}{\mid A\mid+\mid B\mid}
\end{equation}

In medical image segmentation, A usually denotes the algorithm segmentation area, B represents the actual area annotated by experts. The Dice coefficient measures the degree of overlap between these two regions and ranges from 0 to 1, with 0 indicating complete disjointedness and 1 indicating perfect identity between the two sample sets.
}

\subsubsection{NSD}
{
The Normalized Surface Dice offers a refined measure of similarity by accounting for errors at segmentation boundaries \cite{15MedSAm}. Based on standard DSC, NSD emphasizes the overlap between segmentation results and expert-annotated ground truth values specifically at the boundary. The formula is given as:
\begin{equation}\label{eqn-2} 
 NSD=\frac{2\mid S(A)\cap S(B)\mid}{\mid S(A)\mid+\mid S(B)\mid}
\end{equation}

Here, $S(A)$ and $ S(B)$ denote the surface point sets of the region $A$ segmented by the algorithm and the real region $B$ annotated by experts, respectively. $\mid S(A)\mid $ and $\mid S(B)\mid$ represent the number of elements in the surface point sets of $A$ and $B$ ,while $\mid S(A)\cap S(B)\mid$ indicates the number of elements intersecting the surface point sets of  both $A$ and $B$. NSD gives a detailed look at how well a segmentation algorithm works by focusing on the accuracy of the boundary regions between segments. Compared to the standard Dice coefficient, NSD exhibits greater sensitivity to small deviations on the segmentation boundary.
}

\subsubsection{IoU and mIoU}
{
One of the most commonly used measure for the comparison of predicted segmentation mask and the ground truth one is the Intersection over Union (IoU), also called as Jaccard Score \cite{1BCP,12DTC,13MC-Net,27SASSNet,28SS-Net,30UA-MT}. Its formula is defined as:
\begin{equation}\label{eqn-4} 
IoU=\frac{\left|A\cap B\right|}{\left|A\cup B\right|}
\end{equation}

This score corresponds to the ratio of the intersection to the union of two sets, $A$ and $B$. The cardinality of union of sets $A$ and $B$, i.e. the number of elements that falls in either in set $A$ or $B$, is expressed by $\left|A\cup B\right|$, and the cardinality of intersection between sets $A$ and $B$, i.e the number of elements they fall in common, is expressed by $\left|A\cap B\right|$. In medical image segmentation, this metric is used to quantify the similarity between the results of segmentation and the ground-truth of segmentation. The segmentation results and the true labels are considered as two poles, and the bigger the IoU is, the more similar the two shapes overlap with each other, as well the better the segmentation results are. 

For multi-class segmentation, the mean Intersection over Union (mIoU)\cite{53mIoU} is typically used, which computes the average IoU across all classes:
\begin{equation}
\mathrm{mIoU} = \frac{1}{C}\sum_{i=1}^{C}\frac{TP_i}{TP_i + FP_i + FN_i}
\end{equation}

Here, $C$ is the number of classes, and $TP_i$, $FP_i$, and $FN_i$ denote the true positives, false positives, and false negatives of class $i$. The higher mIoU means better segmentation performance for all the classes, making it a reliable metric for globally evaluation in a multi-organ or lesion segmentation task. 
}

\subsubsection{95HD}
{
The 95\% Hausdorff Distance (95HD) is an important metric  derived from Hausdorff distance (HD) to  measure the similarity between two point sets \cite{1BCP,12DTC,13MC-Net,27SASSNet,28SS-Net,30UA-MT}, particularly in medical image segmentation. The Hausdorff Distance measures the greatest discrepancy between two point sets by calculating the maximum value of the minimum distance from each point in one point set to another point set\cite{9CoraNet,25SAMed} . The 95HD is a variant of HD that mitigates the influence of extreme values, resulting in more robust and reliable measurements.

The Hausdorff Distance formula is as follows:
\begin{equation}\label{eqn-5} 
H\left(A,B\right)=\max\left\{h(A,B),h(B,A)\right\}
\end{equation}

Where:
\begin{equation}\label{eqn-6} 
h\left(A,B\right)=\max_{a\in A}\left\{\min_{b\in B}\left\|a-b\right\|\right\}
\end{equation}
\begin{equation}\label{eqn-7} 
h\left(B,A\right)=\max_{b\in B}\left\{\min_{a\in A}\left\|b-a\right\|\right\}
\end{equation}

$h\left(A,B\right)$ and $h\left(B,A\right)$ are referred to as the unidirectional Hausdorff distances from set $A$ to set $B$ and from set $B$ to set $A$, respectively. Specifically, $h\left(A,B\right)$ sorts the distance $\|a_i-b_j\|$ between each point $a_{i}$ in point set $A$ and the nearest point $b_{j}$ in point set $B$ , and then takes the maximum value as $h\left(A,B\right)$ . Similarly, $h\left(B,A\right)$ is obtained, and the maximum value of $h\left(A,B\right)$  and $h\left(B,A\right)$ is taken as the Hausdorff Distance $H\left(A,B\right)$. The 95HD is obtained using the Hausdorff Distance by sorting the distances and taking the 95th percentile of the distance distribution:

\begin{footnotesize}
 	\begin{equation}
 		\begin{aligned}
 			\mathrm{d}_{95}(\mathrm{A},\mathrm{B})=\mathrm{percentile}_{95}(\{d(a,B){:}a\in A\}\cup\{d(b,A){:}b\in B\})
 		\end{aligned}
 		\label{eq_s1}
 	\end{equation}
 \end{footnotesize}

In this case, $d(a,B)$ is the shortest distance from point a to set B, and $d(b,A)$ is the shortest distance from point b to set A. The smaller the 95HD, the closer the segmentation result is to the actual contour, the higher the precision of the segmentation.
}

\subsection{Classification Performance Metrics}

{
For medical image segmentation application, performance is usually assessed using three fundamental  indicators, including Precision, Recall, and F1 Score. Precision\cite{7CM-Global,10DAN,20pionono} measures the ratio of true positive samples to the sum of the true positive and false positive samples, while Recall\cite{10DAN} estimates to which extent the model is able to recognize real positive samples. F1 Score\cite{3APRIL-GAN,10DAN,34Winclip} is the harmonic mean of Precision and Recall and delivers balance evaluation, which is particularly valuable when classes are imbalanced. These metrics are derived from the confusion matrix components, including true positives (TP), false positives (FP), and false negatives (FN) and are basic for measuring model accuracy, sensitivity, and robustness. Together, they offer a concise and interpretable assessment of a model’s classification performance.
}

\subsection{Error Measurement Metrics}
\subsubsection{GED}
{
Generalized Energy Distance is a metric for measuring discrepancies between probability distributions, with particular suitability for assessing the diversity and accuracy of generative models \cite{21Prob.U-Net,36D-Personal,37Tyche}. As an extension and generalization of energy distance, GED captures the differences between predicted and true distributions. It takes into account not only mean and variance, but also the shape of the distribution and other higher-order statistical properties. The formula is given as follows:

\begin{equation}\label{eqn-3} 
\begin{aligned}
&GED=2\mathbb{E}_{(X,Y)\sim P\times Q}\Big[d(X,Y)\Big]\\
& -\mathbb{E}_{(X_1,X_2)\sim P\times P}\Big[d\big(X_1,X_2\big)\Big]\\
&-\mathbb{E}_{(Y_1,Y_2)\sim Q\times Q}\Big[d\big(Y_1,Y_2\big)\Big]
\end{aligned}
\end{equation}

The discrepancies between the ground truth distribution $Q$ and the predicted distribution $P$ are captured by GED. \(P\) and \(Q\) are two probability distributions defined over a common sample space. The variables \(X, X_1, X_2\), and \(Y, Y_1, Y_2\) are independently drawn from \(P\) and \(Q\), respectively. The operator \(\mathbb{E}_{(\cdot,\cdot)\sim A\times B}[\cdot]\) denotes the expectation with respect to all pairs of independently drawn samples from \(A\) and \(B\). The function \(d(\cdot,\cdot)\) is a distance metric that measures dissimilarity between two samples in the space of interest.

GED captures how different the two distributions \(P\) and \(Q\) are by comparing the average distance between cross-distribution pairs \((X, Y)\) with the average distances within each distribution separately. A smaller GED value indicates greater alignment between the predicted and true distributions, implying better predictive performance.
}

\subsubsection{ASSD}
{
The Average Symmetric Surface Distance is an indicator used to assess the performance of medical image segmentation algorithms, quantifying the proximity between segmentation results and ground truth annotations\cite{1BCP,4AutoSAM,12DTC,13MC-Net,27SASSNet,28SS-Net,30UA-MT,32URPC}. Denoting the surface of the segmentation result as $S_{r}$ and the surface of the actual annotation as $S_{t}$, the formula is as follows:

\begin{footnotesize}
 	\begin{equation}
 		\begin{aligned}
 			\mathrm{ASSD}=\frac{1}{N_{r}}\sum_{p_{r\in S_{r}}}min_{p_{t\in S_{t}}}d(p_{r},p_{t})\\+\frac{1}{N_{t}}\sum_{p_{t\in S_{t}}}min_{p_{r\in S_{r}}}d(p_{t},p_{r})
 		\end{aligned}
 		\label{eq_s2}
 	\end{equation}
 \end{footnotesize}

Here, $N_{r}$ and $N_{t}$ represent the number of points on  $S_{r}$ and $S_{t}$, respectively.  $p_{r}$ and $p_{t}$  are such points on $S_{r}$ and $S_{t}$,  while $d(p_{r},p_{t})$ is the Euclidean distance between  $p_{r}$ and  $p_{t}$.  ASSD exhibits symmetry, sensitivity and quantifiable segmentation accuracy. The smaller the value of ASSD, the closer the average distance of the segmentation result surface and the ground truth, and the better is the correspondence and the segmentation accuracy. 
}

\section{Current Major Challenges}

\subsection{Challenges of Data}
\subsubsection{Data distribution mismatch}
{The inconsistency distribution in labeled and unlabeled data is mainly embodied in heterogeneous pathological features, sampling bias, and differences in image quality and acquisition equipment.
This severely hinders the development of medical image segmentation tasks in model training, and consequently leads to low model generalization, training instability, overfitting, forgetting of knowledge from labeled data, and a huge downturn in overall model performance \cite{1BCP,5BGAD,11DRA}. In the future, dynamic distribution matching and continual learning approaches could be explored to adapt to the change of data distribution over time. 
}

\subsubsection{Domain differences }
{A major problem is that natural and medical images are very different on the kind of image content, quality, segmentation targets, and annotation. Although natural imaging processing achieves significant progress, there are distinct particularities in medical images, including high resolution, various anatomical structures, and disease-specific variations, limiting the transferability of natural image models \cite{15MedSAm,23SAM}. Closing this gap necessitates domain-adaptive pre-training and domain-adaptive architectures for biomedical data. 
}

\subsection{Challenges of Model}
\subsubsection{Extremely low contrast}
{Critical anatomical structures and lesions in CT and MRI, such as the edges of tumor, tiny vessels, and mild pathology, are usually with extremely low contrast\cite{28SS-Net}. This leads to blurred boundaries, low signal-to-noise ratios, subtle grayscale differences, and heterogeneous target shapes, complicating segmentation\cite{36D-Personal}.
One way to address it could be uncertainty modeling and attention-based enhancing modules amplifying important low-contrast regions. 
}

\subsubsection{Generality and Adaptability }
{Presently, most models only work on a specific organ or task and still suffer from considerable performance degradation on other datasets or modalities once transferred \cite{4AutoSAM,14Medclip,19Patchcore}.
Although multi-modal large models have emerged, their performance, when compromising for generality, falls far short of the effects achieved by task-specific segmentation models, failing to attain the required capability to adapt to downstream task scenarios. Cross-modal fusion and multi-scale scheme are indispensable to increase the versatility, but how to make complementary information among different modalities well fused is still an open issue \cite{44foundation}. 
Hybrid architectures which utilize the strengths from prompt-based control and module-wise learning might in the future offer improved task adaptability while maintaining generalization. 
}

\subsubsection{Sustainability}
{The laborious and costly data annotation process, \cite{2ADVENT}and the substantial GPU and memory demands for model training \cite{3APRIL-GAN,17MSA}impede the advancement of sustainable segmentation models, contributing to high energy use and carbon emissions. 
Annotation quality and granularity further affect model reliability.
SAM and its variants enhance annotation efficiency via specialized methods, while streamlined model design promises to decrease GPU and memory requirements, fostering the sustainability of AI models. In the future, weak supervision, active learning, and synthetic data generation may become key trends to reduce annotation costs and improve model robustness.}

\subsection{Challenges of  Clinical Applications}
\subsubsection{Annotate ambiguity}
{The ambiguity in the annotations can be attributed mainly to two significant contributors: (i) uncertainties at the data level\cite{36D-Personal}, such as irregular targets, variation in object position, and limited imaging resolution, and (ii) subjectivity at the observer level \cite{7CM-Global,8CM-Pixel,20pionono,29TAB}, where annotators may provide different annotations for the same image due to their personal expertise and experience. The scarcity of experts and inconsistent annotation granularity exacerbate the problem.
Existing segmentation models typically offer only  a single deterministic segmentation outcome, failing to adequately account for these ambiguities and consequently limiting the model's credibility and flexibility in clinical applications.
Incorporating uncertainty quantification, multi-annotator consensus modeling, and probabilistic segmentation could be a potential direction to improve model trustworthiness in the future.}

\subsubsection{The issue of `black box'}
{In medical image segmentation, the ``black box'' problem pertains to the lack of transparency and interpretability in the decision-making processes of deep learning models for clinicians. Despite their good results in segmentation tasks, these models lack understandability in order for a clinician to be able to rely on the prediction made. \cite{30UA-MT,32URPC,35ContrastDiagnosis}. False predictions of the models cannot be analysed to provide an insight on how those errors have occurred and thus has challenged the diagnosis and fixing of these errors. Moreover, due to the requirements for rational reasoning and documentation of diagnostic and therapeutic considerations, the non-interpretability of the model may present legal and ethical issues.
Explainable AI (XAI), visual attention maps, and causal inference-based reasoning are likely to be at the core of solving this.
}

\subsubsection{Simplify visual presentation}
{Information overload, excessive visual complexity and lack of interactivity are some of the challenges that prevent the wide acceptance of the medical image segmentation models on clinics level \cite{24SAMAdapter,26SAM-U,35ContrastDiagnosis}. To achieve intuitive visualization, it is necessary to represent the aggregation of complex segmentation results and the inherent complexity of medical images, so as to ensure that the depth of visualization is compatible with the image complexity.
Moreover, the necessity of segmentation itself is debated: in many scenarios, segmentation cannot serve independently and has to be combined with other diagnostics outputs, implying a trend towards end-to-end models bypassing or performing tight coupling of segmentation with downstream tasks. Nevertheless, accurate 2D and 3D segmentation still has its irreplaceable value in critical applications such as tumor margin delineation, surgical guidance, and intraoperative navigation.
Clinically practical models are likely to be those that offer a high degree of interaction combined with clean visual presentation, allowing clinicians to easily interpret results and enhance diagnostic accuracy.
}

\subsubsection{Timeliness}
{With the evolutionary development and correction of misconceptions in medical field, the correctness of the existing annotations and the generalization ability of the learned model based on these data are limited \cite{45survey}. Temporal validity should be maintained through ongoing learning and updating of the adaptive model within dynamic annotation pipelines, so as to ensure that the models remain clinically relevant.
}

\section{Conclusion}
This article traces the evolution from U-Net to Transformer and then to SAM-based foundation models, emphasizing how segmentation architectures have shifted from task-specific convolutional designs toward globally contextual and prompt-driven paradigms. In the future, hybrid architectures that integrate the merits of these frameworks are promising to boost further improvements on segmentation.
Meanwhile, there is a perspective need to construct clinically-relevant, multi-dimensional evaluation metrics beyond pure accuracy, which have explicit consideration for task relevance, robustness and interpretability. In conclusion, our work provided a thorough and futuristic reference for building accurate generalizable and clinically applicable segmentation models.

\printcredits

\section*{Acknowledgments}
This work was supported by the National Natural Science Foundation of China under Grant No. 61976090 and No. 62476272, and the Fundamental Research Funds for the Central Universities under No. 2025JC001.
\bibliographystyle{cas-model2-names}

\bibliography{main}

\begin{thebibliography}{62}
\expandafter\ifx\csname natexlab\endcsname\relax\def\natexlab#1{#1}\fi
\providecommand{\url}[1]{\texttt{#1}}
\providecommand{\href}[2]{#2}
\providecommand{\path}[1]{#1}
\providecommand{\DOIprefix}{doi:}
\providecommand{\ArXivprefix}{arXiv:}
\providecommand{\URLprefix}{URL: }
\providecommand{\Pubmedprefix}{pmid:}
\providecommand{\doi}[1]{\href{http://dx.doi.org/#1}{\path{#1}}}
\providecommand{\Pubmed}[1]{\href{pmid:#1}{\path{#1}}}
\providecommand{\bibinfo}[2]{#2}
\ifx\xfnm\relax \def\xfnm[#1]{\unskip,\space#1}\fi
\bibitem[{Armato~III et~al.(2011)Armato~III, McLennan, Bidaut, McNitt-Gray,
  Meyer, Reeves, Zhao, Aberle, Henschke, Hoffman et~al.}]{54LIDC-IDRI}
\bibinfo{author}{Armato~III, S.G.}, \bibinfo{author}{McLennan, G.},
  \bibinfo{author}{Bidaut, L.}, \bibinfo{author}{McNitt-Gray, M.F.},
  \bibinfo{author}{Meyer, C.R.}, \bibinfo{author}{Reeves, A.P.},
  \bibinfo{author}{Zhao, B.}, \bibinfo{author}{Aberle, D.R.},
  \bibinfo{author}{Henschke, C.I.}, \bibinfo{author}{Hoffman, E.A.}, et~al.,
  \bibinfo{year}{2011}.
\newblock \bibinfo{title}{The lung image database consortium (lidc) and image
  database resource initiative (idri): a completed reference database of lung
  nodules on ct scans}.
\newblock \bibinfo{journal}{Medical physics} \bibinfo{volume}{38},
  \bibinfo{pages}{915--931}.
\bibitem[{Bai et~al.(2023)Bai, Chen, Li, Shen and Wang}]{1BCP}
\bibinfo{author}{Bai, Y.}, \bibinfo{author}{Chen, D.}, \bibinfo{author}{Li,
  Q.}, \bibinfo{author}{Shen, W.}, \bibinfo{author}{Wang, Y.},
  \bibinfo{year}{2023}.
\newblock \bibinfo{title}{Bidirectional copy-paste for semi-supervised medical
  image segmentation}, in: \bibinfo{booktitle}{Proceedings of the IEEE/CVF
  Conference on Computer Vision and Pattern Recognition}, pp.
  \bibinfo{pages}{11514--11524}.
\bibitem[{Bernard et~al.(2018)Bernard, Lalande, Zotti, Cervenansky, Yang, Heng,
  Cetin, Lekadir, Camara, Ballester et~al.}]{55ACDC}
\bibinfo{author}{Bernard, O.}, \bibinfo{author}{Lalande, A.},
  \bibinfo{author}{Zotti, C.}, \bibinfo{author}{Cervenansky, F.},
  \bibinfo{author}{Yang, X.}, \bibinfo{author}{Heng, P.A.},
  \bibinfo{author}{Cetin, I.}, \bibinfo{author}{Lekadir, K.},
  \bibinfo{author}{Camara, O.}, \bibinfo{author}{Ballester, M.A.G.}, et~al.,
  \bibinfo{year}{2018}.
\newblock \bibinfo{title}{Deep learning techniques for automatic mri cardiac
  multi-structures segmentation and diagnosis: is the problem solved?}
\newblock \bibinfo{journal}{IEEE transactions on medical imaging}
  \bibinfo{volume}{37}, \bibinfo{pages}{2514--2525}.
\bibitem[{Bilic et~al.(2023)Bilic, Christ, Li, Vorontsov, Ben-Cohen, Kaissis,
  Szeskin, Jacobs, Mamani, Chartrand et~al.}]{57lits}
\bibinfo{author}{Bilic, P.}, \bibinfo{author}{Christ, P.}, \bibinfo{author}{Li,
  H.B.}, \bibinfo{author}{Vorontsov, E.}, \bibinfo{author}{Ben-Cohen, A.},
  \bibinfo{author}{Kaissis, G.}, \bibinfo{author}{Szeskin, A.},
  \bibinfo{author}{Jacobs, C.}, \bibinfo{author}{Mamani, G.E.H.},
  \bibinfo{author}{Chartrand, G.}, et~al., \bibinfo{year}{2023}.
\newblock \bibinfo{title}{The liver tumor segmentation benchmark (lits)}.
\newblock \bibinfo{journal}{Medical image analysis} \bibinfo{volume}{84},
  \bibinfo{pages}{102680}.
\bibitem[{Chen et~al.(2023a)Chen, Zhu, Ding, Cao, Zhang, Wang, Li, Sun, Mao and
  Zang}]{24SAMAdapter}
\bibinfo{author}{Chen, T.}, \bibinfo{author}{Zhu, L.}, \bibinfo{author}{Ding,
  C.}, \bibinfo{author}{Cao, R.}, \bibinfo{author}{Zhang, S.},
  \bibinfo{author}{Wang, Y.}, \bibinfo{author}{Li, Z.}, \bibinfo{author}{Sun,
  L.}, \bibinfo{author}{Mao, P.}, \bibinfo{author}{Zang, Y.},
  \bibinfo{year}{2023}a.
\newblock \bibinfo{title}{Sam fails to segment anything?--sam-adapter: Adapting
  sam in underperformed scenes: Camouflage, shadow, and more}.
\newblock \bibinfo{journal}{arXiv preprint arXiv:2304.09148}
  \bibinfo{volume}{2}, \bibinfo{pages}{7}.
\bibitem[{Chen et~al.(2023b)Chen, Han and Zhang}]{3APRIL-GAN}
\bibinfo{author}{Chen, X.}, \bibinfo{author}{Han, Y.}, \bibinfo{author}{Zhang,
  J.}, \bibinfo{year}{2023}b.
\newblock \bibinfo{title}{April-gan: A zero-/few-shot anomaly classification
  and segmentation method for cvpr 2023 vand workshop challenge tracks 1\&2:
  1st place on zero-shot ad and 4th place on few-shot ad}.
\newblock \bibinfo{journal}{arXiv preprint arXiv:2305.17382} .
\bibitem[{Clunie(2001)}]{51dicom}
\bibinfo{author}{Clunie, D.A.}, \bibinfo{year}{2001}.
\newblock \bibinfo{title}{Dicom structured reporting: an object model as an
  implementation boundary}, in: \bibinfo{booktitle}{Medical Imaging 2001: PACS
  and Integrated Medical Information Systems: Design and Evaluation},
  \bibinfo{organization}{SPIE}. pp. \bibinfo{pages}{207--215}.
\bibitem[{Deng et~al.(2023)Deng, Zou, Ren, Wang, Yuan, Ying and Fu}]{26SAM-U}
\bibinfo{author}{Deng, G.}, \bibinfo{author}{Zou, K.}, \bibinfo{author}{Ren,
  K.}, \bibinfo{author}{Wang, M.}, \bibinfo{author}{Yuan, X.},
  \bibinfo{author}{Ying, S.}, \bibinfo{author}{Fu, H.}, \bibinfo{year}{2023}.
\newblock \bibinfo{title}{Sam-u: Multi-box prompts triggered uncertainty
  estimation for reliable sam in medical image}, in:
  \bibinfo{booktitle}{Medical Image Computing and Computer Assisted
  Intervention -- MICCAI 2023 Workshops}, \bibinfo{publisher}{Springer Nature
  Switzerland}, \bibinfo{address}{Cham}. pp. \bibinfo{pages}{368--377}.
\bibitem[{Deng and Li(2022)}]{22RD4AD}
\bibinfo{author}{Deng, H.}, \bibinfo{author}{Li, X.}, \bibinfo{year}{2022}.
\newblock \bibinfo{title}{Anomaly detection via reverse distillation from
  one-class embedding}, in: \bibinfo{booktitle}{Proceedings of the IEEE/CVF
  Conference on Computer Vision and Pattern Recognition}, pp.
  \bibinfo{pages}{9737--9746}.
\bibitem[{Ding et~al.(2022)Ding, Pang and Shen}]{11DRA}
\bibinfo{author}{Ding, C.}, \bibinfo{author}{Pang, G.}, \bibinfo{author}{Shen,
  C.}, \bibinfo{year}{2022}.
\newblock \bibinfo{title}{Catching both gray and black swans: Open-set
  supervised anomaly detection}, in: \bibinfo{booktitle}{Proceedings of the
  IEEE/CVF Conference on Computer Vision and Pattern Recognition}, pp.
  \bibinfo{pages}{7388--7398}.
\bibitem[{Du et~al.(2024)Du, Bai, Huang and Zhao}]{43du2023segvol}
\bibinfo{author}{Du, Y.}, \bibinfo{author}{Bai, F.}, \bibinfo{author}{Huang,
  T.}, \bibinfo{author}{Zhao, B.}, \bibinfo{year}{2024}.
\newblock \bibinfo{title}{Segvol: Universal and interactive volumetric medical
  image segmentation}.
\newblock \bibinfo{journal}{Advances in Neural Information Processing Systems}
  \bibinfo{volume}{37}, \bibinfo{pages}{110746--110783}.
\bibitem[{Gama et~al.(2014)Gama, \v{Z}liobaitundefined, Bifet, Pechenizkiy and
  Bouchachia}]{45survey}
\bibinfo{author}{Gama, J.a.}, \bibinfo{author}{\v{Z}liobaitundefined, I.},
  \bibinfo{author}{Bifet, A.}, \bibinfo{author}{Pechenizkiy, M.},
  \bibinfo{author}{Bouchachia, A.}, \bibinfo{year}{2014}.
\newblock \bibinfo{title}{A survey on concept drift adaptation}.
\newblock \bibinfo{journal}{ACM Comput. Surv.} \bibinfo{volume}{46}.
\newblock \URLprefix \url{https://doi.org/10.1145/2523813},
  \DOIprefix\doi{10.1145/2523813}.
\bibitem[{Gao et~al.(2025)Gao, Liu, Li, Li, Chen, Zhou and
  Metaxas}]{46show2025}
\bibinfo{author}{Gao, Y.}, \bibinfo{author}{Liu, D.}, \bibinfo{author}{Li, Z.},
  \bibinfo{author}{Li, Y.}, \bibinfo{author}{Chen, D.}, \bibinfo{author}{Zhou,
  M.}, \bibinfo{author}{Metaxas, D.N.}, \bibinfo{year}{2025}.
\newblock \bibinfo{title}{Show and segment: Universal medical image
  segmentation via in-context learning}, in: \bibinfo{booktitle}{Proceedings of
  the Computer Vision and Pattern Recognition Conference (CVPR)}, pp.
  \bibinfo{pages}{20830--20840}.
\bibitem[{Gudovskiy et~al.(2022)Gudovskiy, Ishizaka and Kozuka}]{6CFlow-AD}
\bibinfo{author}{Gudovskiy, D.}, \bibinfo{author}{Ishizaka, S.},
  \bibinfo{author}{Kozuka, K.}, \bibinfo{year}{2022}.
\newblock \bibinfo{title}{Cflow-ad: Real-time unsupervised anomaly detection
  with localization via conditional normalizing flows}, in:
  \bibinfo{booktitle}{2022 IEEE/CVF Winter Conference on Applications of
  Computer Vision (WACV)}, pp. \bibinfo{pages}{98--107}.
\bibitem[{He et~al.(2025)He, Huang, Jiang, Nie, Wang, Wang and
  Chen}]{44foundation}
\bibinfo{author}{He, Y.}, \bibinfo{author}{Huang, F.}, \bibinfo{author}{Jiang,
  X.}, \bibinfo{author}{Nie, Y.}, \bibinfo{author}{Wang, M.},
  \bibinfo{author}{Wang, J.}, \bibinfo{author}{Chen, H.}, \bibinfo{year}{2025}.
\newblock \bibinfo{title}{Foundation model for advancing healthcare:
  Challenges, opportunities and future directions}.
\newblock \bibinfo{journal}{IEEE Reviews in Biomedical Engineering}
  \bibinfo{volume}{18}, \bibinfo{pages}{172--191}.
\bibitem[{Heller et~al.(2021)Heller, Isensee, Maier-Hein, Hou, Xie, Li, Nan,
  Mu, Lin, Han, Yao, Gao, Zhang, Wang, Hou, Yang, Xiong, Tian, Zhong, Ma,
  Rickman, Dean, Stai, Tejpaul, Oestreich, Blake, Kaluzniak, Raza, Rosenberg,
  Moore, Walczak, Rengel, Edgerton, Vasdev, Peterson, McSweeney, Peterson,
  Kalapara, Sathianathen, Papanikolopoulos and Weight}]{58kits}
\bibinfo{author}{Heller, N.}, \bibinfo{author}{Isensee, F.},
  \bibinfo{author}{Maier-Hein, K.H.}, \bibinfo{author}{Hou, X.},
  \bibinfo{author}{Xie, C.}, \bibinfo{author}{Li, F.}, \bibinfo{author}{Nan,
  Y.}, \bibinfo{author}{Mu, G.}, \bibinfo{author}{Lin, Z.},
  \bibinfo{author}{Han, M.}, \bibinfo{author}{Yao, G.}, \bibinfo{author}{Gao,
  Y.}, \bibinfo{author}{Zhang, Y.}, \bibinfo{author}{Wang, Y.},
  \bibinfo{author}{Hou, F.}, \bibinfo{author}{Yang, J.},
  \bibinfo{author}{Xiong, G.}, \bibinfo{author}{Tian, J.},
  \bibinfo{author}{Zhong, C.}, \bibinfo{author}{Ma, J.},
  \bibinfo{author}{Rickman, J.}, \bibinfo{author}{Dean, J.},
  \bibinfo{author}{Stai, B.}, \bibinfo{author}{Tejpaul, R.},
  \bibinfo{author}{Oestreich, M.}, \bibinfo{author}{Blake, P.},
  \bibinfo{author}{Kaluzniak, H.}, \bibinfo{author}{Raza, S.},
  \bibinfo{author}{Rosenberg, J.}, \bibinfo{author}{Moore, K.},
  \bibinfo{author}{Walczak, E.}, \bibinfo{author}{Rengel, Z.},
  \bibinfo{author}{Edgerton, Z.}, \bibinfo{author}{Vasdev, R.},
  \bibinfo{author}{Peterson, M.}, \bibinfo{author}{McSweeney, S.},
  \bibinfo{author}{Peterson, S.}, \bibinfo{author}{Kalapara, A.},
  \bibinfo{author}{Sathianathen, N.}, \bibinfo{author}{Papanikolopoulos, N.},
  \bibinfo{author}{Weight, C.}, \bibinfo{year}{2021}.
\newblock \bibinfo{title}{The state of the art in kidney and kidney tumor
  segmentation in contrast-enhanced ct imaging: Results of the kits19
  challenge}.
\newblock \bibinfo{journal}{Medical Image Analysis} \bibinfo{volume}{67},
  \bibinfo{pages}{101821}.
\newblock \URLprefix
  \url{https://www.sciencedirect.com/science/article/pii/S1361841520301857},
  \DOIprefix\doi{https://doi.org/10.1016/j.media.2020.101821}.
\bibitem[{Hu et~al.(2023)Hu, Xu and Shi}]{4AutoSAM}
\bibinfo{author}{Hu, X.}, \bibinfo{author}{Xu, X.}, \bibinfo{author}{Shi, Y.},
  \bibinfo{year}{2023}.
\newblock \bibinfo{title}{How to efficiently adapt large segmentation model
  (sam) to medical images}.
\newblock \bibinfo{journal}{arXiv preprint arXiv:2306.13731} .
\bibitem[{Huang et~al.(2025a)Huang, Zhou, Fu, Zhang, Zhou and Wu}]{62TIP_brats}
\bibinfo{author}{Huang, K.}, \bibinfo{author}{Zhou, T.}, \bibinfo{author}{Fu,
  H.}, \bibinfo{author}{Zhang, Y.}, \bibinfo{author}{Zhou, Y.},
  \bibinfo{author}{Wu, X.J.}, \bibinfo{year}{2025}a.
\newblock \bibinfo{title}{Uncertainty-aware cross-training for semi-supervised
  medical image segmentation}.
\newblock \bibinfo{journal}{IEEE Transactions on Image Processing}
  \bibinfo{volume}{34}, \bibinfo{pages}{5543--5556}.
\newblock \DOIprefix\doi{10.1109/TIP.2025.3599783}.
\bibitem[{Huang et~al.(2025b)Huang, Ge, Liu, Hong, Zhao and Loui}]{60TIP1}
\bibinfo{author}{Huang, S.}, \bibinfo{author}{Ge, Y.}, \bibinfo{author}{Liu,
  D.}, \bibinfo{author}{Hong, M.}, \bibinfo{author}{Zhao, J.},
  \bibinfo{author}{Loui, A.C.}, \bibinfo{year}{2025}b.
\newblock \bibinfo{title}{Rethinking copy-paste for consistency learning in
  medical image segmentation}.
\newblock \bibinfo{journal}{IEEE Transactions on Image Processing}
  \bibinfo{volume}{34}, \bibinfo{pages}{1060--1074}.
\newblock \DOIprefix\doi{10.1109/TIP.2025.3536208}.
\bibitem[{Isensee et~al.(2021)Isensee, Jaeger, Kohl, Petersen and
  Maier-Hein}]{18nnU-Net}
\bibinfo{author}{Isensee, F.}, \bibinfo{author}{Jaeger, P.F.},
  \bibinfo{author}{Kohl, S.A.}, \bibinfo{author}{Petersen, J.},
  \bibinfo{author}{Maier-Hein, K.H.}, \bibinfo{year}{2021}.
\newblock \bibinfo{title}{nnu-net: a self-configuring method for deep
  learning-based biomedical image segmentation}.
\newblock \bibinfo{journal}{Nature methods} \bibinfo{volume}{18},
  \bibinfo{pages}{203--211}.
\bibitem[{Jeong et~al.(2023)Jeong, Zou, Kim, Zhang, Ravichandran and
  Dabeer}]{34Winclip}
\bibinfo{author}{Jeong, J.}, \bibinfo{author}{Zou, Y.}, \bibinfo{author}{Kim,
  T.}, \bibinfo{author}{Zhang, D.}, \bibinfo{author}{Ravichandran, A.},
  \bibinfo{author}{Dabeer, O.}, \bibinfo{year}{2023}.
\newblock \bibinfo{title}{Winclip: Zero-/few-shot anomaly classification and
  segmentation}, in: \bibinfo{booktitle}{Proceedings of the IEEE/CVF Conference
  on Computer Vision and Pattern Recognition}, pp.
  \bibinfo{pages}{19606--19616}.
\bibitem[{Kirillov et~al.(2023)Kirillov, Mintun, Ravi, Mao, Rolland, Gustafson,
  Xiao, Whitehead, Berg, Lo et~al.}]{23SAM}
\bibinfo{author}{Kirillov, A.}, \bibinfo{author}{Mintun, E.},
  \bibinfo{author}{Ravi, N.}, \bibinfo{author}{Mao, H.},
  \bibinfo{author}{Rolland, C.}, \bibinfo{author}{Gustafson, L.},
  \bibinfo{author}{Xiao, T.}, \bibinfo{author}{Whitehead, S.},
  \bibinfo{author}{Berg, A.C.}, \bibinfo{author}{Lo, W.Y.}, et~al.,
  \bibinfo{year}{2023}.
\newblock \bibinfo{title}{Segment anything}, in:
  \bibinfo{booktitle}{Proceedings of the IEEE/CVF International Conference on
  Computer Vision}, pp. \bibinfo{pages}{4015--4026}.
\bibitem[{Kohl et~al.(2018)Kohl, Romera-Paredes, Meyer, De~Fauw, Ledsam,
  Maier-Hein, Eslami, Jimenez~Rezende and Ronneberger}]{21Prob.U-Net}
\bibinfo{author}{Kohl, S.}, \bibinfo{author}{Romera-Paredes, B.},
  \bibinfo{author}{Meyer, C.}, \bibinfo{author}{De~Fauw, J.},
  \bibinfo{author}{Ledsam, J.R.}, \bibinfo{author}{Maier-Hein, K.},
  \bibinfo{author}{Eslami, S.}, \bibinfo{author}{Jimenez~Rezende, D.},
  \bibinfo{author}{Ronneberger, O.}, \bibinfo{year}{2018}.
\newblock \bibinfo{title}{A probabilistic u-net for segmentation of ambiguous
  images}.
\newblock \bibinfo{journal}{Advances in Neural Information Processing Systems}
  \bibinfo{volume}{31}.
\bibitem[{Li et~al.(2024)Li, Navarro, Ezhov, Bayat, Das, Kofler, Shit,
  Waldmannstetter, Paetzold, Hu et~al.}]{59qubiq}
\bibinfo{author}{Li, H.B.}, \bibinfo{author}{Navarro, F.},
  \bibinfo{author}{Ezhov, I.}, \bibinfo{author}{Bayat, A.},
  \bibinfo{author}{Das, D.}, \bibinfo{author}{Kofler, F.},
  \bibinfo{author}{Shit, S.}, \bibinfo{author}{Waldmannstetter, D.},
  \bibinfo{author}{Paetzold, J.C.}, \bibinfo{author}{Hu, X.}, et~al.,
  \bibinfo{year}{2024}.
\newblock \bibinfo{title}{Qubiq: Uncertainty quantification for biomedical
  image segmentation challenge}.
\newblock \bibinfo{journal}{arXiv preprint arXiv:2405.18435} .
\bibitem[{Li et~al.(2020)Li, Zhang and He}]{27SASSNet}
\bibinfo{author}{Li, S.}, \bibinfo{author}{Zhang, C.}, \bibinfo{author}{He,
  X.}, \bibinfo{year}{2020}.
\newblock \bibinfo{title}{Shape-aware semi-supervised 3d semantic segmentation
  for medical images}, in: \bibinfo{booktitle}{Medical Image Computing and
  Computer Assisted Intervention -- MICCAI 2020}, \bibinfo{publisher}{Springer
  International Publishing}, \bibinfo{address}{Cham}. pp.
  \bibinfo{pages}{552--561}.
\bibitem[{Liao et~al.(2025)Liao, Yang, Zhao, Liang and Yuan}]{64TIP_synapse}
\bibinfo{author}{Liao, M.}, \bibinfo{author}{Yang, R.}, \bibinfo{author}{Zhao,
  Y.}, \bibinfo{author}{Liang, W.}, \bibinfo{author}{Yuan, J.},
  \bibinfo{year}{2025}.
\newblock \bibinfo{title}{Focaltransnet: A hybrid focal-enhanced transformer
  network for medical image segmentation}.
\newblock \bibinfo{journal}{IEEE Transactions on Image Processing}
  \bibinfo{volume}{34}, \bibinfo{pages}{5614--5627}.
\newblock \DOIprefix\doi{10.1109/TIP.2025.3602739}.
\bibitem[{Liao et~al.(2023)Liao, Hu, Xie and Xia}]{29TAB}
\bibinfo{author}{Liao, Z.}, \bibinfo{author}{Hu, S.}, \bibinfo{author}{Xie,
  Y.}, \bibinfo{author}{Xia, Y.}, \bibinfo{year}{2023}.
\newblock \bibinfo{title}{Transformer-based annotation bias-aware medical image
  segmentation}, in: \bibinfo{booktitle}{Medical Image Computing and Computer
  Assisted Intervention -- MICCAI 2023}, \bibinfo{publisher}{Springer Nature
  Switzerland}, \bibinfo{address}{Cham}. pp. \bibinfo{pages}{24--34}.
\bibitem[{Luo et~al.(2021a)Luo, Chen, Song and Wang}]{12DTC}
\bibinfo{author}{Luo, X.}, \bibinfo{author}{Chen, J.}, \bibinfo{author}{Song,
  T.}, \bibinfo{author}{Wang, G.}, \bibinfo{year}{2021}a.
\newblock \bibinfo{title}{Semi-supervised medical image segmentation through
  dual-task consistency}, in: \bibinfo{booktitle}{Proceedings of the AAAI
  conference on artificial intelligence}, pp. \bibinfo{pages}{8801--8809}.
\bibitem[{Luo et~al.(2021b)Luo, Liao, Chen, Song, Chen, Zhang, Chen, Wang and
  Zhang}]{32URPC}
\bibinfo{author}{Luo, X.}, \bibinfo{author}{Liao, W.}, \bibinfo{author}{Chen,
  J.}, \bibinfo{author}{Song, T.}, \bibinfo{author}{Chen, Y.},
  \bibinfo{author}{Zhang, S.}, \bibinfo{author}{Chen, N.},
  \bibinfo{author}{Wang, G.}, \bibinfo{author}{Zhang, S.},
  \bibinfo{year}{2021}b.
\newblock \bibinfo{title}{Efficient semi-supervised gross target volume of
  nasopharyngeal carcinoma segmentation via uncertainty rectified pyramid
  consistency}, in: \bibinfo{booktitle}{Medical Image Computing and Computer
  Assisted Intervention -- MICCAI 2021}, \bibinfo{publisher}{Springer
  International Publishing}, \bibinfo{address}{Cham}. pp.
  \bibinfo{pages}{318--329}.
\bibitem[{Ma et~al.(2024)Ma, He, Li, Han, You and Wang}]{15MedSAm}
\bibinfo{author}{Ma, J.}, \bibinfo{author}{He, Y.}, \bibinfo{author}{Li, F.},
  \bibinfo{author}{Han, L.}, \bibinfo{author}{You, C.}, \bibinfo{author}{Wang,
  B.}, \bibinfo{year}{2024}.
\newblock \bibinfo{title}{Segment anything in medical images}.
\newblock \bibinfo{journal}{Nature Communications} \bibinfo{volume}{15},
  \bibinfo{pages}{654}.
\bibitem[{Ma et~al.(2025)Ma, Zhang, Li, Qi, Yu and Shi}]{47steady2025}
\bibinfo{author}{Ma, Q.}, \bibinfo{author}{Zhang, J.}, \bibinfo{author}{Li,
  Z.}, \bibinfo{author}{Qi, L.}, \bibinfo{author}{Yu, Q.},
  \bibinfo{author}{Shi, Y.}, \bibinfo{year}{2025}.
\newblock \bibinfo{title}{Steady progress beats stagnation: Mutual aid of
  foundation and conventional models in mixed domain semi-supervised medical
  image segmentation}, in: \bibinfo{booktitle}{Proceedings of the Computer
  Vision and Pattern Recognition Conference}, pp. \bibinfo{pages}{5175--5185}.
\bibitem[{Menze et~al.(2015)Menze, Jakab, Bauer, Kalpathy-Cramer, Farahani,
  Kirby, Burren, Porz, Slotboom, Wiest, Lanczi, Gerstner, Weber, Arbel, Avants,
  Ayache, Buendia, Collins, Cordier, Corso, Criminisi, Das, Delingette,
  Demiralp, Durst, Dojat, Doyle, Festa, Forbes, Geremia, Glocker, Golland, Guo,
  Hamamci, Iftekharuddin, Jena, John, Konukoglu, Lashkari, Mariz, Meier,
  Pereira, Precup, Price, Raviv, Reza, Ryan, Sarikaya, Schwartz, Shin, Shotton,
  Silva, Sousa, Subbanna, Szekely, Taylor, Thomas, Tustison, Unal, Vasseur,
  Wintermark, Ye, Zhao, Zhao, Zikic, Prastawa, Reyes and Van~Leemput}]{49BraTS}
\bibinfo{author}{Menze, B.H.}, \bibinfo{author}{Jakab, A.},
  \bibinfo{author}{Bauer, S.}, \bibinfo{author}{Kalpathy-Cramer, J.},
  \bibinfo{author}{Farahani, K.}, \bibinfo{author}{Kirby, J.},
  \bibinfo{author}{Burren, Y.}, \bibinfo{author}{Porz, N.},
  \bibinfo{author}{Slotboom, J.}, \bibinfo{author}{Wiest, R.},
  \bibinfo{author}{Lanczi, L.}, \bibinfo{author}{Gerstner, E.},
  \bibinfo{author}{Weber, M.A.}, \bibinfo{author}{Arbel, T.},
  \bibinfo{author}{Avants, B.B.}, \bibinfo{author}{Ayache, N.},
  \bibinfo{author}{Buendia, P.}, \bibinfo{author}{Collins, D.L.},
  \bibinfo{author}{Cordier, N.}, \bibinfo{author}{Corso, J.J.},
  \bibinfo{author}{Criminisi, A.}, \bibinfo{author}{Das, T.},
  \bibinfo{author}{Delingette, H.}, \bibinfo{author}{Demiralp, {\c{C}}.},
  \bibinfo{author}{Durst, C.R.}, \bibinfo{author}{Dojat, M.},
  \bibinfo{author}{Doyle, S.}, \bibinfo{author}{Festa, J.},
  \bibinfo{author}{Forbes, F.}, \bibinfo{author}{Geremia, E.},
  \bibinfo{author}{Glocker, B.}, \bibinfo{author}{Golland, P.},
  \bibinfo{author}{Guo, X.}, \bibinfo{author}{Hamamci, A.},
  \bibinfo{author}{Iftekharuddin, K.M.}, \bibinfo{author}{Jena, R.},
  \bibinfo{author}{John, N.M.}, \bibinfo{author}{Konukoglu, E.},
  \bibinfo{author}{Lashkari, D.}, \bibinfo{author}{Mariz, J.A.},
  \bibinfo{author}{Meier, R.}, \bibinfo{author}{Pereira, S.},
  \bibinfo{author}{Precup, D.}, \bibinfo{author}{Price, S.J.},
  \bibinfo{author}{Raviv, T.R.}, \bibinfo{author}{Reza, S.M.S.},
  \bibinfo{author}{Ryan, M.}, \bibinfo{author}{Sarikaya, D.},
  \bibinfo{author}{Schwartz, L.}, \bibinfo{author}{Shin, H.C.},
  \bibinfo{author}{Shotton, J.}, \bibinfo{author}{Silva, C.A.},
  \bibinfo{author}{Sousa, N.}, \bibinfo{author}{Subbanna, N.K.},
  \bibinfo{author}{Szekely, G.}, \bibinfo{author}{Taylor, T.J.},
  \bibinfo{author}{Thomas, O.M.}, \bibinfo{author}{Tustison, N.J.},
  \bibinfo{author}{Unal, G.}, \bibinfo{author}{Vasseur, F.},
  \bibinfo{author}{Wintermark, M.}, \bibinfo{author}{Ye, D.H.},
  \bibinfo{author}{Zhao, L.}, \bibinfo{author}{Zhao, B.},
  \bibinfo{author}{Zikic, D.}, \bibinfo{author}{Prastawa, M.},
  \bibinfo{author}{Reyes, M.}, \bibinfo{author}{Van~Leemput, K.},
  \bibinfo{year}{2015}.
\newblock \bibinfo{title}{The multimodal brain tumor image segmentation
  benchmark (brats)}.
\newblock \bibinfo{journal}{IEEE Transactions on Medical Imaging}
  \bibinfo{volume}{34}, \bibinfo{pages}{1993--2024}.
\newblock \DOIprefix\doi{10.1109/TMI.2014.2377694}.
\bibitem[{Milletari et~al.(2016)Milletari, Navab and Ahmadi}]{33V-Net}
\bibinfo{author}{Milletari, F.}, \bibinfo{author}{Navab, N.},
  \bibinfo{author}{Ahmadi, S.A.}, \bibinfo{year}{2016}.
\newblock \bibinfo{title}{V-net: Fully convolutional neural networks for
  volumetric medical image segmentation}, in: \bibinfo{booktitle}{2016 Fourth
  International Conference on 3D Vision (3DV)}, pp. \bibinfo{pages}{565--571}.
\bibitem[{Qin et~al.(2025)Qin, Liu, Zhang, Li, Guo and Guo}]{63TIP_DSC}
\bibinfo{author}{Qin, G.}, \bibinfo{author}{Liu, H.}, \bibinfo{author}{Zhang,
  X.}, \bibinfo{author}{Li, W.}, \bibinfo{author}{Guo, Y.},
  \bibinfo{author}{Guo, C.}, \bibinfo{year}{2025}.
\newblock \bibinfo{title}{Semi-supervised medical hyperspectral image
  segmentation using adversarial consistency constraint learning and cross
  indication network}.
\newblock \bibinfo{journal}{IEEE Transactions on Image Processing}
  \bibinfo{volume}{34}, \bibinfo{pages}{5414--5428}.
\newblock \DOIprefix\doi{10.1109/TIP.2025.3598499}.
\bibitem[{Qiu et~al.(2025)Qiu, Zhang and Li}]{61TIP_acdc}
\bibinfo{author}{Qiu, C.H.}, \bibinfo{author}{Zhang, X.S.},
  \bibinfo{author}{Li, Y.J.}, \bibinfo{year}{2025}.
\newblock \bibinfo{title}{Joanet: An integrated joint optimization architecture
  making medical image segmentation really helped by super-resolution
  pre-processing}.
\newblock \bibinfo{journal}{IEEE Transactions on Image Processing} ,
  \bibinfo{pages}{1--1}\DOIprefix\doi{10.1109/TIP.2025.3620627}.
\bibitem[{Rajchl et~al.(2017)Rajchl, Lee, Oktay, Kamnitsas, Passerat-Palmbach,
  Bai, Damodaram, Rutherford, Hajnal, Kainz and Rueckert}]{42by2}
\bibinfo{author}{Rajchl, M.}, \bibinfo{author}{Lee, M.C.H.},
  \bibinfo{author}{Oktay, O.}, \bibinfo{author}{Kamnitsas, K.},
  \bibinfo{author}{Passerat-Palmbach, J.}, \bibinfo{author}{Bai, W.},
  \bibinfo{author}{Damodaram, M.}, \bibinfo{author}{Rutherford, M.A.},
  \bibinfo{author}{Hajnal, J.V.}, \bibinfo{author}{Kainz, B.},
  \bibinfo{author}{Rueckert, D.}, \bibinfo{year}{2017}.
\newblock \bibinfo{title}{Deepcut: Object segmentation from bounding box
  annotations using convolutional neural networks}.
\newblock \bibinfo{journal}{IEEE Transactions on Medical Imaging}
  \bibinfo{volume}{36}, \bibinfo{pages}{674--683}.
\bibitem[{Rakic et~al.(2024)Rakic, Wong, Ortiz, Cimini, Guttag and
  Dalca}]{37Tyche}
\bibinfo{author}{Rakic, M.}, \bibinfo{author}{Wong, H.E.},
  \bibinfo{author}{Ortiz, J.J.G.}, \bibinfo{author}{Cimini, B.A.},
  \bibinfo{author}{Guttag, J.V.}, \bibinfo{author}{Dalca, A.V.},
  \bibinfo{year}{2024}.
\newblock \bibinfo{title}{Tyche: Stochastic in-context learning for medical
  image segmentation}, in: \bibinfo{booktitle}{Proceedings of the IEEE/CVF
  Conference on Computer Vision and Pattern Recognition}, pp.
  \bibinfo{pages}{11159--11173}.
\bibitem[{Rayed et~al.(2024)Rayed, Islam, Niha, Jim, Kabir and Mridha}]{40re3}
\bibinfo{author}{Rayed, M.E.}, \bibinfo{author}{Islam, S.S.},
  \bibinfo{author}{Niha, S.I.}, \bibinfo{author}{Jim, J.R.},
  \bibinfo{author}{Kabir, M.M.}, \bibinfo{author}{Mridha, M.},
  \bibinfo{year}{2024}.
\newblock \bibinfo{title}{Deep learning for medical image segmentation:
  State-of-the-art advancements and challenges}.
\newblock \bibinfo{journal}{Informatics in Medicine Unlocked}
  \bibinfo{volume}{47}, \bibinfo{pages}{101504}.
\newblock \URLprefix
  \url{https://www.sciencedirect.com/science/article/pii/S2352914824000601},
  \DOIprefix\doi{https://doi.org/10.1016/j.imu.2024.101504}.
\bibitem[{Ronneberger et~al.(2015)Ronneberger, Fischer and Brox}]{31U-Net}
\bibinfo{author}{Ronneberger, O.}, \bibinfo{author}{Fischer, P.},
  \bibinfo{author}{Brox, T.}, \bibinfo{year}{2015}.
\newblock \bibinfo{title}{U-net: Convolutional networks for biomedical image
  segmentation}, in: \bibinfo{booktitle}{Medical Image Computing and
  Computer-Assisted Intervention -- MICCAI 2015}, \bibinfo{publisher}{Springer
  International Publishing}, \bibinfo{address}{Cham}. pp.
  \bibinfo{pages}{234--241}.
\bibitem[{Roth et~al.(2022)Roth, Pemula, Zepeda, Sch{\"o}lkopf, Brox and
  Gehler}]{19Patchcore}
\bibinfo{author}{Roth, K.}, \bibinfo{author}{Pemula, L.},
  \bibinfo{author}{Zepeda, J.}, \bibinfo{author}{Sch{\"o}lkopf, B.},
  \bibinfo{author}{Brox, T.}, \bibinfo{author}{Gehler, P.},
  \bibinfo{year}{2022}.
\newblock \bibinfo{title}{Towards total recall in industrial anomaly
  detection}, in: \bibinfo{booktitle}{Proceedings of the IEEE/CVF Conference on
  Computer Vision and Pattern Recognition}, pp. \bibinfo{pages}{14318--14328}.
\bibitem[{Salehi et~al.(2021)Salehi, Sadjadi, Baselizadeh, Rohban and
  Rabiee}]{16MKD}
\bibinfo{author}{Salehi, M.}, \bibinfo{author}{Sadjadi, N.},
  \bibinfo{author}{Baselizadeh, S.}, \bibinfo{author}{Rohban, M.H.},
  \bibinfo{author}{Rabiee, H.R.}, \bibinfo{year}{2021}.
\newblock \bibinfo{title}{Multiresolution knowledge distillation for anomaly
  detection}, in: \bibinfo{booktitle}{Proceedings of the IEEE/CVF Conference on
  Computer Vision and Pattern Recognition}, pp. \bibinfo{pages}{14902--14912}.
\bibitem[{Salpea et~al.(2023)Salpea, Tzouveli and Kollias}]{53mIoU}
\bibinfo{author}{Salpea, N.}, \bibinfo{author}{Tzouveli, P.},
  \bibinfo{author}{Kollias, D.}, \bibinfo{year}{2023}.
\newblock \bibinfo{title}{Medical image segmentation: A review of modern
  architectures}, in: \bibinfo{booktitle}{Computer Vision -- ECCV 2022
  Workshops}, \bibinfo{publisher}{Springer Nature Switzerland},
  \bibinfo{address}{Cham}. pp. \bibinfo{pages}{691--708}.
\bibitem[{Sati et~al.(2012)Sati, George, Shea, Gait\'{a}n and Reich}]{50flair}
\bibinfo{author}{Sati, P.}, \bibinfo{author}{George, I.C.},
  \bibinfo{author}{Shea, C.D.}, \bibinfo{author}{Gait\'{a}n, M.I.},
  \bibinfo{author}{Reich, D.S.}, \bibinfo{year}{2012}.
\newblock \bibinfo{title}{Flair*: A combined mr contrast technique for
  visualizing white matter lesions and parenchymal veins}.
\newblock \bibinfo{journal}{Radiology} \bibinfo{volume}{265},
  \bibinfo{pages}{926--932}.
\newblock \URLprefix \url{https://doi.org/10.1148/radiol.12120208},
  \DOIprefix\doi{10.1148/radiol.12120208},
  \href{http://arxiv.org/abs/https://doi.org/10.1148/radiol.12120208}{\tt
  arXiv:https://doi.org/10.1148/radiol.12120208}. \bibinfo{note}{pMID:
  23074257}.
\bibitem[{Schmidt et~al.(2023)Schmidt, Morales-Alvarez and Molina}]{20pionono}
\bibinfo{author}{Schmidt, A.}, \bibinfo{author}{Morales-Alvarez, P.},
  \bibinfo{author}{Molina, R.}, \bibinfo{year}{2023}.
\newblock \bibinfo{title}{Probabilistic modeling of inter-and intra-observer
  variability in medical image segmentation}, in:
  \bibinfo{booktitle}{Proceedings of the IEEE/CVF International Conference on
  Computer Vision}, pp. \bibinfo{pages}{21097--21106}.
\bibitem[{Shi et~al.(2024)Shi, Ma, Yang, Wang and Zhang}]{41by1}
\bibinfo{author}{Shi, Y.}, \bibinfo{author}{Ma, J.}, \bibinfo{author}{Yang,
  J.}, \bibinfo{author}{Wang, S.}, \bibinfo{author}{Zhang, Y.},
  \bibinfo{year}{2024}.
\newblock \bibinfo{title}{Beyond pixel-wise supervision for medical image
  segmentation: From traditional models to foundation models}.
\newblock \bibinfo{journal}{arXiv preprint arXiv:2404.13239} .
\bibitem[{Shi et~al.(2021)Shi, Zhang, Ling, Lu, Zheng, Yu, Qi and
  Gao}]{9CoraNet}
\bibinfo{author}{Shi, Y.}, \bibinfo{author}{Zhang, J.}, \bibinfo{author}{Ling,
  T.}, \bibinfo{author}{Lu, J.}, \bibinfo{author}{Zheng, Y.},
  \bibinfo{author}{Yu, Q.}, \bibinfo{author}{Qi, L.}, \bibinfo{author}{Gao,
  Y.}, \bibinfo{year}{2021}.
\newblock \bibinfo{title}{Inconsistency-aware uncertainty estimation for
  semi-supervised medical image segmentation}.
\newblock \bibinfo{journal}{IEEE transactions on medical imaging}
  \bibinfo{volume}{41}, \bibinfo{pages}{608--620}.
\bibitem[{Tanno et~al.(2019)Tanno, Saeedi, Sankaranarayanan, Alexander and
  Silberman}]{7CM-Global}
\bibinfo{author}{Tanno, R.}, \bibinfo{author}{Saeedi, A.},
  \bibinfo{author}{Sankaranarayanan, S.}, \bibinfo{author}{Alexander, D.C.},
  \bibinfo{author}{Silberman, N.}, \bibinfo{year}{2019}.
\newblock \bibinfo{title}{Learning from noisy labels by regularized estimation
  of annotator confusion}, in: \bibinfo{booktitle}{Proceedings of the IEEE/CVF
  Conference on Computer Vision and Pattern Recognition}, pp.
  \bibinfo{pages}{11244--11253}.
\bibitem[{Vu et~al.(2019)Vu, Jain, Bucher, Cord and P{\'e}rez}]{2ADVENT}
\bibinfo{author}{Vu, T.H.}, \bibinfo{author}{Jain, H.},
  \bibinfo{author}{Bucher, M.}, \bibinfo{author}{Cord, M.},
  \bibinfo{author}{P{\'e}rez, P.}, \bibinfo{year}{2019}.
\newblock \bibinfo{title}{Advent: Adversarial entropy minimization for domain
  adaptation in semantic segmentation}, in: \bibinfo{booktitle}{Proceedings of
  the IEEE/CVF Conference on Computer Vision and Pattern Recognition}, pp.
  \bibinfo{pages}{2517--2526}.
\bibitem[{Wang et~al.(2024)Wang, Yi, Wang, Zhang, Liu, Mori, Yuan and
  Yang}]{35ContrastDiagnosis}
\bibinfo{author}{Wang, C.}, \bibinfo{author}{Yi, Y.}, \bibinfo{author}{Wang,
  Y.}, \bibinfo{author}{Zhang, C.}, \bibinfo{author}{Liu, Y.},
  \bibinfo{author}{Mori, K.}, \bibinfo{author}{Yuan, M.},
  \bibinfo{author}{Yang, G.}, \bibinfo{year}{2024}.
\newblock \bibinfo{title}{Contrastdiagnosis: Enhancing interpretability in lung
  nodule diagnosis using contrastive learning}.
\newblock \bibinfo{journal}{arXiv preprint arXiv:2403.05280} .
\bibitem[{Wang et~al.(2022)Wang, Wu, Agarwal and Sun}]{14Medclip}
\bibinfo{author}{Wang, Z.}, \bibinfo{author}{Wu, Z.}, \bibinfo{author}{Agarwal,
  D.}, \bibinfo{author}{Sun, J.}, \bibinfo{year}{2022}.
\newblock \bibinfo{title}{Medclip: Contrastive learning from unpaired medical
  images and text}, in: \bibinfo{booktitle}{Proceedings of the Conference on
  Empirical Methods in Natural Language Processing.}, p. \bibinfo{pages}{3876}.
\bibitem[{Wu et~al.(2025)Wu, Wang, Hong, Ji, Fu, Xu, Xu and Jin}]{17MSA}
\bibinfo{author}{Wu, J.}, \bibinfo{author}{Wang, Z.}, \bibinfo{author}{Hong,
  M.}, \bibinfo{author}{Ji, W.}, \bibinfo{author}{Fu, H.}, \bibinfo{author}{Xu,
  Y.}, \bibinfo{author}{Xu, M.}, \bibinfo{author}{Jin, Y.},
  \bibinfo{year}{2025}.
\newblock \bibinfo{title}{Medical sam adapter: Adapting segment anything model
  for medical image segmentation}.
\newblock \bibinfo{journal}{Medical Image Analysis} \bibinfo{volume}{102},
  \bibinfo{pages}{103547}.
\newblock \URLprefix
  \url{https://www.sciencedirect.com/science/article/pii/S1361841525000945},
  \DOIprefix\doi{https://doi.org/10.1016/j.media.2025.103547}.
\bibitem[{Wu et~al.(2024)Wu, Luo, Xu, Guo, Ju, Ge, Liao and Cai}]{36D-Personal}
\bibinfo{author}{Wu, Y.}, \bibinfo{author}{Luo, X.}, \bibinfo{author}{Xu, Z.},
  \bibinfo{author}{Guo, X.}, \bibinfo{author}{Ju, L.}, \bibinfo{author}{Ge,
  Z.}, \bibinfo{author}{Liao, W.}, \bibinfo{author}{Cai, J.},
  \bibinfo{year}{2024}.
\newblock \bibinfo{title}{Diversified and personalized multi-rater medical
  image segmentation}, in: \bibinfo{booktitle}{Proceedings of the IEEE/CVF
  Conference on Computer Vision and Pattern Recognition}, pp.
  \bibinfo{pages}{11470--11479}.
\bibitem[{Wu et~al.(2022)Wu, Wu, Wu, Ge and Cai}]{28SS-Net}
\bibinfo{author}{Wu, Y.}, \bibinfo{author}{Wu, Z.}, \bibinfo{author}{Wu, Q.},
  \bibinfo{author}{Ge, Z.}, \bibinfo{author}{Cai, J.}, \bibinfo{year}{2022}.
\newblock \bibinfo{title}{Exploring smoothness and class-separation for
  semi-supervised medical image segmentation}, in: \bibinfo{booktitle}{Medical
  Image Computing and Computer Assisted Intervention -- MICCAI 2022},
  \bibinfo{publisher}{Springer Nature Switzerland}, \bibinfo{address}{Cham}.
  pp. \bibinfo{pages}{34--43}.
\bibitem[{Wu et~al.(2021)Wu, Xu, Ge, Cai and Zhang}]{13MC-Net}
\bibinfo{author}{Wu, Y.}, \bibinfo{author}{Xu, M.}, \bibinfo{author}{Ge, Z.},
  \bibinfo{author}{Cai, J.}, \bibinfo{author}{Zhang, L.}, \bibinfo{year}{2021}.
\newblock \bibinfo{title}{Semi-supervised left atrium segmentation with mutual
  consistency training}, in: \bibinfo{booktitle}{Medical Image Computing and
  Computer Assisted Intervention -- MICCAI 2021}, \bibinfo{publisher}{Springer
  International Publishing}, \bibinfo{address}{Cham}. pp.
  \bibinfo{pages}{297--306}.
\bibitem[{Xiong et~al.(2021)Xiong, Xia, Hu, Huang, Bian, Zheng, Vesal,
  Ravikumar, Maier, Yang, Heng, Ni, Li, Tong, Si, Puybareau, Khoudli, Géraud,
  Chen, Bai, Rueckert, Xu, Zhuang, Luo, Jia, Sermesant, Liu, Wang, Borra,
  Masci, Corsi, {de Vente}, Veta, Karim, Preetha, Engelhardt, Qiao, Wang, Tao,
  Nuñez-Garcia, Camara, Savioli, Lamata and Zhao}]{56LA}
\bibinfo{author}{Xiong, Z.}, \bibinfo{author}{Xia, Q.}, \bibinfo{author}{Hu,
  Z.}, \bibinfo{author}{Huang, N.}, \bibinfo{author}{Bian, C.},
  \bibinfo{author}{Zheng, Y.}, \bibinfo{author}{Vesal, S.},
  \bibinfo{author}{Ravikumar, N.}, \bibinfo{author}{Maier, A.},
  \bibinfo{author}{Yang, X.}, \bibinfo{author}{Heng, P.A.},
  \bibinfo{author}{Ni, D.}, \bibinfo{author}{Li, C.}, \bibinfo{author}{Tong,
  Q.}, \bibinfo{author}{Si, W.}, \bibinfo{author}{Puybareau, E.},
  \bibinfo{author}{Khoudli, Y.}, \bibinfo{author}{Géraud, T.},
  \bibinfo{author}{Chen, C.}, \bibinfo{author}{Bai, W.},
  \bibinfo{author}{Rueckert, D.}, \bibinfo{author}{Xu, L.},
  \bibinfo{author}{Zhuang, X.}, \bibinfo{author}{Luo, X.},
  \bibinfo{author}{Jia, S.}, \bibinfo{author}{Sermesant, M.},
  \bibinfo{author}{Liu, Y.}, \bibinfo{author}{Wang, K.},
  \bibinfo{author}{Borra, D.}, \bibinfo{author}{Masci, A.},
  \bibinfo{author}{Corsi, C.}, \bibinfo{author}{{de Vente}, C.},
  \bibinfo{author}{Veta, M.}, \bibinfo{author}{Karim, R.},
  \bibinfo{author}{Preetha, C.J.}, \bibinfo{author}{Engelhardt, S.},
  \bibinfo{author}{Qiao, M.}, \bibinfo{author}{Wang, Y.}, \bibinfo{author}{Tao,
  Q.}, \bibinfo{author}{Nuñez-Garcia, M.}, \bibinfo{author}{Camara, O.},
  \bibinfo{author}{Savioli, N.}, \bibinfo{author}{Lamata, P.},
  \bibinfo{author}{Zhao, J.}, \bibinfo{year}{2021}.
\newblock \bibinfo{title}{A global benchmark of algorithms for segmenting the
  left atrium from late gadolinium-enhanced cardiac magnetic resonance
  imaging}.
\newblock \bibinfo{journal}{Medical Image Analysis} \bibinfo{volume}{67},
  \bibinfo{pages}{101832}.
\newblock \URLprefix
  \url{https://www.sciencedirect.com/science/article/pii/S1361841520301961},
  \DOIprefix\doi{https://doi.org/10.1016/j.media.2020.101832}.
\bibitem[{Yao et~al.(2024)Yao, Bai, Liao, Chen, Liu and Xie}]{39re2}
\bibinfo{author}{Yao, W.}, \bibinfo{author}{Bai, J.}, \bibinfo{author}{Liao,
  W.}, \bibinfo{author}{Chen, Y.}, \bibinfo{author}{Liu, M.},
  \bibinfo{author}{Xie, Y.}, \bibinfo{year}{2024}.
\newblock \bibinfo{title}{From cnn to transformer: A review of medical image
  segmentation models}.
\newblock \bibinfo{journal}{Journal of Imaging Informatics in Medicine}
  \bibinfo{volume}{37}, \bibinfo{pages}{1529--1547}.
\bibitem[{Yao et~al.(2023)Yao, Li, Zhang, Sun and Zhang}]{5BGAD}
\bibinfo{author}{Yao, X.}, \bibinfo{author}{Li, R.}, \bibinfo{author}{Zhang,
  J.}, \bibinfo{author}{Sun, J.}, \bibinfo{author}{Zhang, C.},
  \bibinfo{year}{2023}.
\newblock \bibinfo{title}{Explicit boundary guided semi-push-pull contrastive
  learning for supervised anomaly detection}, in:
  \bibinfo{booktitle}{Proceedings of the IEEE/CVF Conference on Computer Vision
  and Pattern Recognition}, pp. \bibinfo{pages}{24490--24499}.
\bibitem[{Yu et~al.(2019)Yu, Wang, Li, Fu and Heng}]{30UA-MT}
\bibinfo{author}{Yu, L.}, \bibinfo{author}{Wang, S.}, \bibinfo{author}{Li, X.},
  \bibinfo{author}{Fu, C.W.}, \bibinfo{author}{Heng, P.A.},
  \bibinfo{year}{2019}.
\newblock \bibinfo{title}{Uncertainty-aware self-ensembling model for
  semi-supervised 3d left atrium segmentation}, in: \bibinfo{booktitle}{Medical
  Image Computing and Computer Assisted Intervention -- MICCAI 2019},
  \bibinfo{publisher}{Springer International Publishing},
  \bibinfo{address}{Cham}. pp. \bibinfo{pages}{605--613}.
\bibitem[{Zhang and Liu(2023)}]{25SAMed}
\bibinfo{author}{Zhang, K.}, \bibinfo{author}{Liu, D.}, \bibinfo{year}{2023}.
\newblock \bibinfo{title}{Customized segment anything model for medical image
  segmentation}.
\newblock \bibinfo{journal}{arXiv preprint arXiv:2304.13785} .
\bibitem[{Zhang et~al.(2020)Zhang, Tanno, Xu, Jin, Jacob, Cicarrelli, Barkhof
  and Alexander}]{8CM-Pixel}
\bibinfo{author}{Zhang, L.}, \bibinfo{author}{Tanno, R.}, \bibinfo{author}{Xu,
  M.C.}, \bibinfo{author}{Jin, C.}, \bibinfo{author}{Jacob, J.},
  \bibinfo{author}{Cicarrelli, O.}, \bibinfo{author}{Barkhof, F.},
  \bibinfo{author}{Alexander, D.}, \bibinfo{year}{2020}.
\newblock \bibinfo{title}{Disentangling human error from ground truth in
  segmentation of medical images}.
\newblock \bibinfo{journal}{Advances in Neural Information Processing Systems}
  \bibinfo{volume}{33}, \bibinfo{pages}{15750--15762}.
\bibitem[{Zhang et~al.(2024)Zhang, Shen and Jiao}]{38re1}
\bibinfo{author}{Zhang, Y.}, \bibinfo{author}{Shen, Z.}, \bibinfo{author}{Jiao,
  R.}, \bibinfo{year}{2024}.
\newblock \bibinfo{title}{Segment anything model for medical image
  segmentation: Current applications and future directions}.
\newblock \bibinfo{journal}{Computers in Biology and Medicine}
  \bibinfo{volume}{171}, \bibinfo{pages}{108238}.
\newblock \URLprefix
  \url{https://www.sciencedirect.com/science/article/pii/S0010482524003226},
  \DOIprefix\doi{https://doi.org/10.1016/j.compbiomed.2024.108238}.
\bibitem[{Zhang et~al.(2017)Zhang, Yang, Chen, Fredericksen, Hughes and
  Chen}]{10DAN}
\bibinfo{author}{Zhang, Y.}, \bibinfo{author}{Yang, L.}, \bibinfo{author}{Chen,
  J.}, \bibinfo{author}{Fredericksen, M.}, \bibinfo{author}{Hughes, D.P.},
  \bibinfo{author}{Chen, D.Z.}, \bibinfo{year}{2017}.
\newblock \bibinfo{title}{Deep adversarial networks for biomedical image
  segmentation utilizing unannotated images}, in: \bibinfo{booktitle}{Medical
  Image Computing and Computer Assisted Intervention -- MICCAI 2017},
  \bibinfo{publisher}{Springer International Publishing},
  \bibinfo{address}{Cham}. pp. \bibinfo{pages}{408--416}.

\end{thebibliography}


\end{document}